%% file: arxiv.tex
\documentclass{article}

\usepackage{arxiv}
\usepackage[utf8]{inputenc}
\usepackage{latexsym}
\usepackage{graphicx}
\usepackage{verbatim}
\usepackage{times}

\usepackage{amsmath}
\usepackage{tikz}
\usepackage{amsfonts}
\usepackage{subfig}
\usepackage{color}
\usepackage{blindtext}
\usepackage{wrapfig}
\usepackage{tabularx}
\usepackage{multirow}
\usepackage{booktabs}
\usepackage{siunitx}
\usepackage{wasysym}
\usepackage{textcomp}     
\usepackage{setspace}   

\usetikzlibrary{external}

\newcommand{\dd}[1]{\mathrm{d}#1}

\newcommand{\norm}[2]{\left\lVert#1\right\rVert_{#2}}
\newcommand{\mean}[1]{\left\langle #1\right\rangle}
\newcommand{\std}[1]{\sigma\left(#1\right)}

\DeclareSIUnit{\year}{yr}
 
\DeclareMathOperator*{\argmin}{arg\,min} 

\title{Real-Time Optimal Guidance and Control for Interplanetary Transfers Using Deep Networks}

\author{%
Dario Izzo,%
\thanks{Scientific coordinator, Advanced Concepts Team}
Ekin \"{O}zt\"{u}rk%
\thanks{Young Graduate Trainee, Advanced Concepts Team}
}
\begin{document}
\maketitle
\doublespacing

\begin{abstract}
We consider the Earth-Venus mass-optimal interplanetary transfer of a low-thrust spacecraft and show how the optimal guidance can be represented by deep networks in a large portion of the state space and to a high degree of accuracy. Imitation (supervised) learning of optimal examples is used as a network training paradigm. The resulting models are suitable for an on-board, real-time, implementation of the optimal guidance and control system of the spacecraft and are called G\&CNETs. A new general methodology called \lq\lq Backward Generation of Optimal Examples\rq\rq\ is introduced and shown to be able to efficiently create all the optimal state action pairs necessary to train G\&CNETs without solving optimal control problems. With respect to previous works, we are able to produce datasets  containing a few orders of magnitude more optimal trajectories and obtain network performances compatible with real missions requirements. Several schemes able to train representations of either the optimal policy (thrust profile) or the value function (optimal mass) are proposed and tested. We find that both policy learning and value function learning successfully and accurately learn the optimal thrust and that a spacecraft employing the learned thrust is able to reach the target conditions orbit spending only 2\permil\ more propellant than in the corresponding mathematically optimal transfer. Moreover, the optimal propellant mass can be predicted (in case of value function learning) within an error well within 1\%. All G\&CNETs produced are tested during simulations of interplanetary transfers with respect to their ability to reach the target conditions optimally starting from nominal and off-nominal conditions.
\end{abstract}

\maketitle

\section*{Nomenclature}
\label{sec:nomenclature}
\begin{center}
\begin{minipage}{0.48\columnwidth}
\begin{tabular}{@{}lcl@{}}
\textit{{$\mathbf x$}}          &=&    modified equinoctial elements \\
\textit{$\mathbf r, \mathbf v$}          &=&     spacecraft position and velocity \\
\textit{$a$}          &=&     semimajor axis, AU \\
\textit{$e$}          &=&     eccentricity \\
\textit{$i$}          &=&     inclination, rad \\
\textit{$\omega$}          &=&     argument of perigee, rad \\
\textit{$\Omega$}          &=&     right ascension of the \\
&&ascending node, rad \\
\textit{$\nu$}          &=&     true anomaly, rad \\
\textit{p}          &=&     semilatus rectum \\
\textit{f}          &=&     equinoctial parameter \\
\textit{g}          &=&     equinoctial parameter \\
\textit{h}          &=&     equinoctial parameter \\
\textit{k}          &=&     equinoctial parameter \\
\textit{L}          &=&     true longitude \\
\textit{m}          &=&     spacecraft mass, kg\\
\textit{t}          &=&     time, s \\
\textit{{$v$}}          &=&     value function \\
\textit{$\boldsymbol \lambda$}   &=&     equinoctial costate vector \\
\textit{$\mathbf u$}       &=&     control variables\\
\textit{$f_t$}       &=&     radial component of thrust force, N \\
\textit{$f_r$}       &=&     tangential component of thrust force, N \\
\textit{$f_n$}       &=&     normal component of thrust force, N \\
\end{tabular} 
\end{minipage}
\hfill
\begin{minipage}{0.48\columnwidth}
\begin{tabular}{@{}lcl@{}}
\textit{T}          &=&     time scale,  s\\
\textit{A}          &=&     acceleration scale,  m/s${}^2$\\
\textit{{$g_0$}}          &=&     acceleration at ground level, m/s${}^2$ \\
\textit{{$\mathcal T$}}          &=& Set of all optimal trajectories\\
&&for a given dynamics and cost function    \\
\textit{$\mathcal D$}          &=&    database of optimal (augmented) \\
&& state-action pairs\\
\textit{{$T_{\text{max}}$}}          &=&     maximum thrust, N \\
\textit{{$I_{\text{sp}}$}}       &=&     specific impulse, s\\
\textit{{$t_{\text f}$}}       &=&     time of flight\\
\textit{$J$}          &=&     cost function \\
\textit{$\mathcal H$}   &=&     Hamiltonian \\
\textit{$\lambda_0$}          &=&     Hamiltonian scale \\
\textit{$\epsilon$}          &=&     continuation parameter \\
\textit{$\delta$}          &=&     perturbation scale \\
\textit{$c_1$}          &=&     Maximum thrust, N \\
\textit{$c_2$}          &=&     Ratio between $c_1$ and $I_{sp}g_0$, kg/s \\
\textit{$c$}          &=&     Sundmann transform scale \\
\textit{$n$}          &=&     Sundmann transform order \\
\textit{$\mathcal{N}$}   &=&     an artificial neural network \\
\textit{$g$ }              &=&     activation function of a neural network \\
\textit{$l$ }              &=&     loss function of a neural network \\
\textit{rEd}              &=&     reduced Euclidean distance (distance \\
&& between the first 5 equinoctial elements) \\
\end{tabular}
\end{minipage}
\end{center}

\vspace{5mm}
\begin{center}
\textit{Subscripts and Superscripts}

\begin{minipage}{0.48\columnwidth}

\begin{tabular}{@{}lcl@{}}
\textit{$(\cdot)_{\mathcal{N}}$}   &=& neural network prediction 
\end{tabular} 
\end{minipage}
\begin{minipage}{0.48\columnwidth}
\begin{tabular}{@{}lcl@{}}
\textit{$(\cdot)^*$} &=& optimal value
\end{tabular}
\end{minipage}
\end{center}

\setcounter{table}{0}

\section{Introduction}
\label{sec:intro}
The use of deep networks in decision making systems has produced increasingly interesting results over the past decade and in diverse applications ranging from computer gaming to robotics, to micro and unmanned air vehicles and spacecraft \cite{izzo2018survey}.
The mathematical theories powering most of these new results are the consolidated ones of reinforcement learning, dynamic programming and optimal control, coupled to emerging results in training deep networks.
It is worth noting that environments such as those encountered in video-games, Earth robotics and air vehicles are characterized by high noise levels and unpredictability. In all these cases, decision making is greatly affected by environmental stochasticity, making the use of optimal control theory for deterministic systems less appealing.
Spacecraft, on the other hand, operate in a rather different environment, comparatively free of major disturbances. As a consequence, deterministic optimal control methods (e.g. indirect methods based on Pontryagin's principle \cite{pontryagin} or direct methods based on collocation \cite{betts}) are widely used to design guidance profiles for low-thrust interplanetary transfers, spacecraft landing, docking problems etc.
The works from Sanchez and Izzo \cite{sanchez2016learning, sanchez2018real} introduced the idea to use imitation learning (also known as behavioural cloning, and essentially based on the classical supervised learning scheme) to teach a deep artificial neural network to produce on-board, and in real time, the optimal guidance and tested it on several spacecraft landing scenarios.
The results, triggering a number of other studies \cite{cheng2020real, cheng2018real, li2019neural, li2019autonomous, izzo2018stability, tailor2019learning})  suggest that future space systems might use an artificial neural network in place of their on-board guidance and control systems, and hence these networks are called G\&CNETs.
An early study on the stability of a G\&CNET controlled system \cite{izzo2018stability} shows how it is also possible to provide control guarantees to the resulting neurocontrolled system, a fact of great relevance for such a mission critical component.
The extension of these results to interplanetary low-thrust trajectories seems ripe.
In particular it is of interest to prove that G\&CNETs can be developed to represent the highly discontinuous optimal controls arising in mass optimal interplanetary transfers, and in a large portion of the interplanetary medium. 
While not directly using the term G\&CNET, a first study on deep networks for the real time optimal control of interplanetary transfers appeared recently \cite{cheng2018real}, but only considering two dimensional dynamics and a simple solar sailing transfer with continuous controls. 
In following works from Li et al. \cite{li2019neural, li2019autonomous} neural networks are also trained to approximate the co-states, the optimal thrust and the value function of optimal interplanetary transfers, but only succeeding for time optimal cases (resulting in continuous thrust profiles) and in close neighbourhoods of nominal transfers (e.g. small perturbations of the order of 0.1 m/s on the initial velocity and 100m on the initial position were considered \cite{li2019autonomous}). 
The time consuming creation of optimal examples, likely prevented these works to produce large enough datasets and thus train networks able to go well beyond the simple cases considered there and able to approximate the optimal guidance in a larger portion of interplanetary space with acceptable accuracies. 
In a previous, yet preliminary, work \cite{izzo2019gecco} we hinted on how to take advantage of Pontryagin principle to create massive datasets of optimal trajectories avoiding the time consuming solution procedures of optimal control problems.
In this work, we refine those results studying in depth the methodologies there only sketched. Our final aim is to prove the possibility to design G\&CNETs able to produce complex mass optimal guidance profiles in large portions of interplanetary space (i.e. also far away from a nominal transfer). 
We introduce a new generic methodology (the \lq\lq backward generation of optimal examples\rq\rq), based on Pontryagin principle and able to create optimal training samples by numerically integrating a system of equations. We apply it to an Earth-Venus transfer assembling seven large databases of optimal low-thrust transfers which we release publicly (see \cite{ozturk_ekin_2020_3613772} and similar). 
Overall, in the context of this work, we are able to compute and release 4,000,000 mass optimal transfers containing multiple bang-off phases. 
For comparison, the datasets used in \cite{li2019neural} contain roughly 12,000 trajectories, while the datasets used in \cite{cheng2018real, li2019autonomous} contain 1,000 trajectories (and all considering smooth thrust profiles).
After the dataset creation, four different training methodologies are studied: one based on policy learning (i.e. the original G\&CNET training method developed in \cite{sanchez2018real, tailor2019learning}) and three new training procedures based on value function learning (i.e. learning the final optimal propellant mass and inferring the optimal thrust profile from it).
Our paper is structured as follows: in Section \ref{sec:background} we provide the necessary mathematical definitions of the low-thrust interplanetary dynamics considered, the related optimal control problem and we present a short discussion on the consequences of Pontryagin's and Bellman's optimality principles to our case.
In Section \ref{sec:datageneration} we describe the \lq\lq backward generation of optimal examples\rq\rq, a new methodology (based on Pontryagin's principle) to generate large databases of optimal state-action pairs without solving optimal control problems.
Then, in Section \ref{sec:network_training_framework}, the artificial neural network architectures and various loss functions are discussed.
In the following Section \ref{sec:database_generation} we describe the details of the seven different databases of optimal interplanetary transfers to Venus that we use in this paper and that we created and made publicly available (see \cite{ozturk_ekin_2020_3613772}).
Details on the G\&CNET training are then given in the following Section \ref{sec:training}, while the evaluation of the their performances is discussed at length in Section \ref{sec:results}. 

\section{Background}
\label{sec:background}
\subsection{Dynamics}
We consider the motion of a spacecraft of mass $m$ with a position $\mathbf{r}$ and velocity $\mathbf{v}$ subject only to the Sun gravitational attraction in the heliocentric International Celestial Reference Frame (ICRF). The spacecraft also has an ion thruster with a specific impulse $I_{sp}$ and a maximum thrust $c_1$ independent from solar distance.
We describe the spacecraft state via its mass $m$ and the modified equinoctial elements $\mathbf{x}=\left[p,f,g,h,k,L\right]^T$ as originally defined by Walker et al.~\cite{walker}.

The motion of the spacecraft is described by the following set of differential equations:
\begin{equation}
\begin{array}{l}
\dot p = \sqrt{\frac p\mu} \frac{2p}{w} f_t \\
\dot f = \frac{1}{m} \sqrt{\frac p\mu} \left\{f_r \sin L + \left[ (1+w)\cos L + f \right] \frac{f_t}{w} - (h\sin L-k\cos L)\frac{g\cdot f_n}{w} \right\} \\
\dot g = \frac{1}{m} \sqrt{\frac p\mu} \left\{ - f_r\cos L + \left[ (1+w)\sin L + g \right] \frac{f_t}{w} + (h\sin L-k\cos L)\frac{f\cdot f_n}{w} \right\} \\
\dot h = \sqrt{\frac p\mu} \frac{s^2f_n}{2mw}\cos L \\
\dot k = \sqrt{\frac p\mu} \frac{s^2f_n}{2mw}\sin L \\
\dot L = \sqrt{\frac p\mu}\left\{\mu\left(\frac wp\right)^2 + \frac 1w\left(h\sin L-k\cos L\right) \frac{f_n}{m}\right\} \\
\dot m = - \frac{\sqrt{f_r^2+f_t^2+f_n^2}}{I_{sp}g_0}
\end{array}
\end{equation}
\normalsize
where, $w = 1 + f\cos L + g\sin L$, $s^2 = 1 + h^2 + k^2$ and $f_r, f_t, f_n$ are the radial, tangential and normal components of the force generated by the ion thruster. The gravity parameter is denoted with $\mu$ and the gravitational acceleration at sea level with $g_0$.

It is useful to rewrite these equations using matrix notation thus we introduce the matrices $\mathbf{B}$ and $\mathbf{D}$ defined as:

\begin{equation}
\sqrt{\frac \mu p} \mathbf B(\mathbf x) = 
\left[
\begin{array}{ccc}
0 & \frac {2p}w & 0 \\
 \sin L & [(1+w)\cos L + f]\frac 1w  & - \frac gw (h\sin L-k\cos L)  \\
- \cos L & [(1+w)\sin L + g]\frac 1w  & \frac fw (h\sin L-k\cos L) \\
0 & 0  & \frac 1w \frac{s^2}{2}\cos L \\
0 & 0  & \frac 1w \frac{s^2}{2}\sin L \\
0 & 0  & \frac 1w (h\sin L - k\cos L) \\
\end{array}
\right]
\end{equation}
and
\begin{equation}
\mathbf D(\mathbf x) = 
\left[
\begin{array}{cccccc}
0 & 0 & 0 & 0 & 0 & \sqrt{\frac{\mu}{p^3}}w^2
\end{array}
\right]^T
\end{equation}
and $\mathbf x = [p,f,g,h,k,L]^T$. Thus the equations of motion become:
\begin{equation}
\left\{
\begin{array}{l}
\dot {\mathbf x} =  \frac {c_1 u(t)} m \mathbf B(\mathbf x)  \mathbf{\hat i_\tau}  + \mathbf D(\mathbf x) \\
\dot m = -c_2 u(t)
\end{array}
\right.
\label{eq:eom}
\end{equation}
where the spacecraft thrust is now indicated by $c_1 u \mathbf{\hat i_\tau} = [f_r, f_t, f_n]^T$ and is bounded by the relations $|u(t)| \le 1$ and $|\mathbf{\hat i_\tau}(t)| = 1$. The control $u$ is called the throttle and, unlike the thrust, is non dimensional. The dynamics is controlled, at each instant, by the throttle magnitude $u(t) \in [0,1]$ and the thrust direction $\mathbf{\hat i_\tau}$. We refer to these control variables also with a single symbol $\mathbf u(t) = [u(t), \mathbf{\hat i_\tau(t)}]$ and we use the notation $\mathbf u\in \mathcal U$ to indicate that the control $\mathbf u$ belongs to the space $\mathcal U$ of admissible controls. Recall that the constant $c_1$ is the maximum thrust and note that the constant $c_2 = c_1 / (I_{sp}\ g_0)$ was introduced for convenience.

\subsection{The optimal control problem}
\label{sec:problem}
We consider here a free time orbital transfer problem, i.e. finding the controls $u(t)$ and $\mathbf{\hat i_\tau}(t)$ defined in $[0, t_f]$ and the transfer time $t_f$ so that the functional:
\begin{equation}
    \label{eq:cost}
    J(u(t), t_f) = \int_{0}^{t_f}\left\{u - \epsilon \log{[u(1-u)]} \right\} \dd{t}
\end{equation}
is minimized, and the spacecraft is steered from its initial mass $m_0$ and some initial point $\mathbf x_0$ to some final mass $m_f$ and some final point $\mathbf x_f$. The functional $J$, chosen following the work of Bertrand and Epenoy~\cite{bertrand2002new}, is parameterised by a continuation parameter $\epsilon \in [0,1]$ which activates a logarithmic barrier smoothing the problem and ensuring that the constraint $u\in[0,1]$ is always satisfied. Clearly, as the continuation parameter approaches zero $\epsilon\rightarrow 0$ the functional becomes $J = (m_0 - m_f)/c_2$ (substitute Eq.~\eqref{eq:eom} into Eq.~\eqref{eq:cost}), and the problem considered becomes equivalent to minimising the propellant mass.

\subsection{Consequences of Pontryagin's Minimum Principle}
\label{sec:pontryagin}

Following the work of Pontryagin~\cite{pontryagin}, we infer the necessary conditions for optimality by applying Pontryagin's minimum principle. Since we have stated a minimisation problem, the conditions are slightly different from the ones originally derived in Pontryagin's work.
First we introduce the co-states $\boldsymbol\lambda, \mathbf\lambda_m$ as continuous functions defined in $[0, t_f]$ and define the Hamiltonian:
\begin{equation}
\label{eq:hamil}
    \mathcal H(\mathbf x, m, \boldsymbol\lambda, \lambda_m, \mathbf u)  = \frac {c_1 u} m \boldsymbol\lambda^T \mathbf B(\mathbf x) \mathbf{\hat i_\tau}  +  \lambda_L \sqrt{\frac \mu {p^3}}w^2 - c_2 \lambda_m u + \left\{ u - \epsilon \log[u(1-u)] \right\}
\end{equation}
and the system of equations:
\begin{equation}
\label{eq:eom_costates}
\left\{
\begin{array}{l}
\dot {\mathbf x} =  \frac{\partial \mathcal H}{\partial \boldsymbol\lambda}=  \frac {c_1 u(t)} m \mathbf B \mathbf{\hat i_\tau}(t) + \mathbf D  \\
\dot m =  \frac{\partial \mathcal H}{\partial \lambda_m} = -c_2 u(t)  \\
\dot{ \boldsymbol\lambda} = - \frac{\partial \mathcal H}{\partial \mathbf x} \\
\dot{\lambda}_m = -\frac{\partial \mathcal H}{\partial m} \\
\end{array}
\right.
\end{equation}
The explicit form of the various derivatives appearing in the equations above is reported in Appendix~\ref{sec:app}.
Along an optimal trajectory, the Hamiltonian must be zero (free terminal time problem) and minimal with respect to the choices of $u$ and $\mathbf{\hat i_\tau}$. For the optimal thrust direction $\mathbf{\hat i_\tau}^*$ it follows that
\begin{equation}
\label{eq:optimal_direction}
\mathbf{\hat i_\tau}  = \mathbf{\hat i_\tau}^*(t) = - \frac{ \mathbf B^T\boldsymbol\lambda}{| \mathbf B^T\boldsymbol\lambda|}
\end{equation}
where the time dependence on the right hand side is present both in the co-states and in $\mathbf B$, but has been omitted for brevity. For the optimal throttle $u^*$, necessarily:
\begin{equation}
\label{eq:optimal_magnitude}
u(t) = u^*(t) = \frac{2\epsilon}{2\epsilon + SF(t) + \sqrt{4\epsilon^2 + SF(t)^2}}
\end{equation}
where we introduced a switching function:
\begin{equation}
\label{eq:switching}
SF(t) = 1 - \frac{c_1}{m }| \mathbf B^T\boldsymbol\lambda|-c_2\lambda_m.
\end{equation}

\subsubsection{The two point boundary value problem}
\label{sec:tpbvp}
Substituting Eq.~\eqref{eq:hamil}, Eq.~\eqref{eq:optimal_direction} and Eq.~\eqref{eq:optimal_magnitude} into Eq.~\eqref{eq:eom_costates} one obtains a set of ordinary differential equations in the augmented state $(\mathbf x, m, \boldsymbol \lambda, \lambda_m)$ whose solutions represent a generic optimal interplanetary transfer (under the considered dynamics and merit function). We indicate the set of all  solutions to those equations with $\mathcal T$. Typically, one is only interested in searching in $\mathcal T$ for a solution that satisfies the initial conditions on the spacecraft state and some added conditions dictated by Pontryagin's theory (transversality and free-time conditions). Let us consider the case of a transfer from any $\mathbf x_0, m_0$ to Venus orbit (not a rendezvous): the final values for the mass and the true longitude $L$ are left free (transversality conditions, $\lambda_{L}\rvert_{t=t_f} = 0, \lambda_{m}\rvert_{t=t_f} = 0$) while the value of the Hamiltonian is fixed (free-time condition $\mathcal H\rvert_{t=t_f}=0$). Hence we search in $\mathcal T$ for a solution where the initial values over $\mathbf x$ and $m$ are known as well as the final values over $p,f,g,h,k,\lambda_L, \lambda_m, \mathcal H$. This creates a two-boundary value problem that can be solved by a shooting method. In other words, for a given initial state $\mathbf{x_0}$ and $m_0$, we need to find the initial values $\boldsymbol{\lambda}_0$ and $\lambda_{m_0}$ such that solving the initial value problem (IVP) for Eq.~\eqref{eq:eom_costates} results in a final state at $t_f$ that matches the arrival conditions at the target orbit, the transversality conditions and the free-time condition on the Hamiltonian. Formally, we introduce the shooting function:

\begin{equation}
    \label{eq:shooting}
    \phi(\boldsymbol \lambda_0, \lambda_{m_0}, t_f) = \left[p_f-p_V, f_f-f_V, g_f-g_V, h_f-h_V, k_f-k_V, \lambda_{L_f}, \lambda_{m_f}, \mathcal H_f\right]
\end{equation}
where we indicate with a subscript $V$ the modified equinoctial elements of Venus orbit and with the subscript $f$ the final values of the modified equinoctial elements resulting from numerically integrating  Eq.~\eqref{eq:eom_costates} for a time $t_f$ from the initial conditions $\mathbf x_0, m_0, \boldsymbol  \lambda_0, \lambda_{m_0}$. Solving an instance of the optimal control problem considered is then equivalent to solving the equation: $\phi(\boldsymbol \lambda_0, \lambda_{m_0}, t_f)  = 0$.

\subsection{Consequences of Bellman's Principle of Optimality}
\label{sec:bellman}
Let us now apply Bellman's principle of optimality~\cite{bellman1966dynamic} to the optimal control problem we stated in Section \ref{sec:problem}. We indicate with $v(\mathbf x, m)$ the value function, i.e. the optimal value of the functional defined by Eq.~\eqref{eq:cost} when the initial spacecraft state is $\mathbf x, m$. Since the value function is, in the case considered, time-independent, the Hamilton Jacobi Bellman (HJB) equations can be written as:

\begin{eqnarray}
 0 =  \min_{\mathbf u \in \mathcal U}(u + \nabla_{\mathbf x} v \cdot \mathbf f(\mathbf x, m)) \label{eq:HJB1} \\
\mathbf u = \argmin_{\mathbf u \in \mathcal U}(u + \nabla_{\mathbf x} v \cdot \mathbf f(\mathbf x, m))  \label{eq:HJB2}.
\end{eqnarray}
These equations hold for all points where $v(\mathbf x, m)$ is differentiable. We use $\mathbf f(\mathbf x, m)$ to denote the right hand side of Eq.~\eqref{eq:eom} including the mass equation and, abusing the notation, we use, only in this context, the symbol $\boldsymbol \lambda$ to indicate all the co-states.
Pontryagin's minimum principle can then, in general, be formulated as $\mathbf u = \argmin_{\mathbf u\in \mathcal U}\mathcal H = \argmin_{\mathbf u\in \mathcal U}\left(u + \boldsymbol \lambda \cdot \mathbf f(\mathbf x, m)\right)$. Comparing this last expression to Eq.(\eqref{eq:HJB2}) we may conclude that the co-states introduced by Pontryagin in his theory are the gradients of the value function introduced by Bellman in his theory. This fact, albeit rarely exploited in interplanetary trajectory optimization research, provides a convenient basis for the design of artificial neural network learning procedures, a fact we exploit later in Section \ref{sec:training}.

\section{Generating Databases of Optimal Trajectories}
\label{sec:datageneration}

As the main purpose of this work is to study artificial neural networks capability to learn the structure of optimal low-thrust interplanetary trajectories, a learning database  $\mathcal D:=\{(\mathbf x, m, \boldsymbol \lambda, \lambda_m), \mathbf u^*_i) .. i=1...N\}$ containing spacecraft (augmented) states and the optimal associated thrust vectors is needed. In this section we describe a method able to efficiently construct such a database. In order to generate $\mathcal D$, the straight forward approach would be to solve a large number of optimal control problems, sample each resulting optimal trajectory in multiple time instants and store the resultant (augmented) state-action pairs. Such an approach has indeed been pursued in the past, also by us (CITE), but it comes with an intrinsic problem: finding one optimal interplanetary low-thrust trajectory is a computationally intensive task, finding thousands of them can be a barrier limiting the applicability of the resulting method. 
An alternative method, called \lq\lq Backward Generation of Optimal Examples\rq\rq\ is here presented that is able to create a sufficiently dense database of optimal trajectories by solving only once the Two-Point Boundary Value Problem (TPBVP) resulting from the application of Pontryagin principle, and then exploiting principles of optimality to generate more optimal trajectories to learn from at the cost of simpler numerical integrations. A good statistical distribution of the sampled data points is then also ensured by making use of the Sundman~\cite{Sundman1913} transform during such integrations.

\subsection {Backward Generation of Optimal Examples}
\label{sec:computationcost}
In this section we describe our method to construct efficiently a database $\mathcal D$ of optimal state action pairs: the \lq\lq Backward Generation of Optimal Examples\rq\rq. The brute-force approach for populating $\mathcal D$ requires selecting a number of meaningful initial states and then solving the corresponding optimal control problem (e.g. finding a zero for the shooting function, see Section \ref{sec:background}). This approach scales extremely poorly because of the known computational difficulties associated with solving optimal control problems, both using direct and indirect methods. Previous work (e.g. Sanchez and Izzo~\cite{sanchez2016learning, sanchez2018real}) deployed a continuation (homotopy) approach to reduce some of the complexity involved by eliminating the need to search for a new initial guess for each optimal control problem instance: by perturbing the initial state of a nominal trajectory, the unperturbed states (and co-states) provide (most of the time) a good initial guess to solve also the newly created optimal control problem. However, assuming that no convergence issues occur, it is still necessary to solve Eq.~\eqref{eq:shooting} for the new initial conditions considered (if an indirect method is pursued) or the resulting transcribed non linear programming problem (if a direct method is pursued). In both cases a significant computational cost is encountered.

In practice, there exists a more efficient way to obtain a similar result avoiding entirely convergence issues, guaranteeing optimality and reducing the computational costs significantly. The idea, applicable more generally to any problem formulated in the form introduced in Section \ref{sec:tpbvp}, is to perform a backward in time numerical propagation of Eq.~\eqref{eq:eom_costates} starting from suitable values of the state and co-states obtained perturbing the final values known for a nominal trajectory. This perturbation needs to be chosen such that the transversality conditions and the condition on the Hamiltonian are still satisfied. If this is the case, the trajectory obtained by the backward integration of Eq.~\eqref{eq:eom_costates} will be the solution to the optimal control problem of reaching the target orbit from any state along the computed trajectory. Hence we can insert any point along such a trajectory into our database $\mathcal D$.

Formally, consider a nominal optimal trajectory and indicate with $\mathbf x_f^*, m_f^*, \boldsymbol \lambda_f^*, \lambda_{m_f}^*$ the final values (i.e. at $t^*_f$) of the states and the co-states. Consider then a new set of co-state values:
\begin{equation}
\boldsymbol \lambda^{new}_f = \boldsymbol \lambda_f^* + \delta \boldsymbol \lambda, \quad
\lambda^{new}_{m_f} = \lambda_{m_f}^* + \delta \lambda_{m_f}
\end{equation}
where the perturbation $\delta \boldsymbol \lambda$ is chosen in some ball $B_\rho \in \mathbb R^7$ of size $\rho$. 
Since we want that the transversality conditions on the final (free) true longitude and on the final (free) mass to be satisfied also for the new costates, we set $\delta \lambda_{L} = 0,\ \delta \lambda_{m_f} = 0$. The remaining values for $\delta \boldsymbol \lambda$ are randomly sampled within $B_\rho$. 
If the values for the new states $\mathbf x^{new}_f, m^{new}_f$ were now kept equal to the nominal ones, the trajectory resulting from propagating backward in time Eq.~\eqref{eq:eom_costates} from new final conditions $\mathbf x_f^{new}, m_f^{new}, \boldsymbol \lambda^{new}_f, \lambda^{new}_{m_f}$ would be fulfilling all of Pontryagin's necessary conditions for optimality, except $\mathcal H_f = 0$, and would be arriving to the target orbit with the same mass and true longitude $L$ as the nominal trajectory. 
Thus we perturb also the two states $m^{new}_f = m_f^* + \delta m$ and $L^{new}_f = L_f^* + \delta L$, first choosing $\delta m$ at random (within some bounds $\rho$) and then considering the Hamiltonian as a function of the sole anomaly perturbation, $\mathcal H_f(\delta L) = 0$, and solving for $\delta L$. The resulting trajectory, integrated backward from the new conditions, can be sampled and inserted into $\mathcal D$. Such a trajectory neither ends where the nominal trajectory ends, nor starts from where the nominal trajectory starts, but it is nevertheless optimal (with respect to Eq.~\eqref{eq:cost}) and represents a valid interplanetary transfer to learn from (it reaches the target orbit).
this technique is what we call \lq\lq Backward Generation of Optimal Examples\rq\rq. It reduces the cost of computing one more optimal trajectory to that of a backward in time integration of Eq.~\eqref{eq:eom_costates} plus the computational cost of a single root finding call to solve $\mathcal H_f(\delta L) = 0$.

\subsection{Sampling the optimal trajectories}
\label{sec:sundman_transform}
Each optimal trajectory obtained, regardless of the method used to compute it, has to be sampled in $N$ points and the corresponding (augmented) state-action pairs inserted into $\mathcal D$. In order to create a database covering equally the interplanetary space we avoid the use of sampling the optimal trajectories uniformly in time (which would create a strong bias for larger distances) and use, instead, the Sundman transformation originally described by Sundman~\cite{Sundman1913} and Levi-Civita~\cite{LeviCivita1906}. The integration variable in Eq.~\eqref{eq:eom_costates} is thus transformed from $\dd{t}$ to $\dd{\theta_s}$:
\begin{equation}
\label{eq:sundman}
    \dd{t}=cr^n\dd{\theta_s}
\end{equation}
where $c$ is a constant that depends on $n$ and $r$ is the radial distance from the main body. In this work, we use $n=1$ and $c=\sqrt{a/\mu}$ where $\theta_s$ is, therefore, the eccentric anomaly and $a$ is the semi-major axis. We use the notation $\dot p$ to refer to derivatives with respect to time and $p^\prime$ to refer to derivatives with respect to $\theta_s$. Using the Sundman transformation, Eq.~\eqref{eq:eom_costates} becomes:
\begin{equation}
\label{eq:eom_costates_sund}
\left\{
\begin{array}{l}
{\mathbf x}^\prime = \dot{x}\sqrt{\frac{a(\theta_s)}{\mu}} =  \left[\frac {c_1 u(\theta_s)} m \mathbf B \mathbf{\hat i_\tau}(\theta_s) + \mathbf D\right]\sqrt{\frac{a(\theta_s)}{\mu}}  \\
m^\prime = \dot{m}\sqrt{\frac{a(\theta_s)}{\mu}} = \left[-c_2 u(\theta_s)\right]\sqrt{\frac{a(\theta_s)}{\mu}}  \\
{\mathbf\lambda}^\prime = \left[\dot{\mathbf{\lambda}}\right]\sqrt{\frac{a(\theta_s)}{\mu}} \\
{\lambda}^\prime_m = \left[\dot{\lambda}_m\right]\sqrt{\frac{a(\theta_s)}{\mu}} \\
t^\prime = \sqrt{\frac{a(\theta_s)}{\mu}} \\
\end{array}
\right.
\end{equation}
When implementing the discussed \lq\lq Backward Generation of Optimal Examples\rq\rq\ methodology, we use this system of equations, rather than Eq.~\eqref{eq:eom_costates}, to perform the backward propagation. The resulting trajectory is then sampled at $N$ equally spaced points in $\theta_s$, which results in a sample density uniform in the eccentric anomaly. Note that varying the parameter $n$ of the Sundmann transformation one can bias the database $\mathcal D$, and thus the network learning, to have different sample densities at different radial distances. The choice made in this work is motivated by the idea to not give preference to positions far away from the central point of attraction.
\input{gecnet_draw.tex}

\section{The Network}
\label{sec:network_training_framework}

In this section we describe the neural network architectures we use and the various loss functions proposed and studied for their training. The final aim of training an artificial neural network on $\mathcal D$ is to have it learn the optimal control structure for a low-thrust transfer and thus use it on-board the satellite to generate, in real time and autonomously, the guidance and control commands to be sent to the spacecraft thrusters. For this reason, although the term originated in a different context, these type of networks are referred to as Guidance and Control Networks or, briefly, G\&CNETs \cite{izzo2018stability}.

In this study we indicate, generically, a G\&CNET with the symbol $\mathcal N(\mathbf x, m)$. Formally this is a function of the spacecraft state, defined by the following relations:
\begin{equation}
\label{eq:ann}
\mathcal N(\mathbf x, m): \enskip
\left\{
\begin{array}{l}
L^{(0)}  [\mathbf x, m] \\
L^{(i+1)} = \sigma_i(\boldsymbol W^{(i)} L^{(i)} + \mathbf b^{(i)}), \enskip \forall i = 0..l \\
\mathcal N(\mathbf x, m) = L^{(l+1)}
\end{array}
\right.
\end{equation}
where $L^{(0)}$ denotes the input layer and $L^{(i+1)}$ the subsequent hidden layers.
The network \textit{depth} (i.e. the number of layers) $l$  as well as the network \textit{width} (i.e. the number of neurons per layers) determined by the weight matrices $\boldsymbol W^{(i)}$ and bias vectors $\mathbf b^{(i)}$ dimensions, constitute the network architecture.
$\sigma_i$ is a non-linear function termed \textit{activation function} 
that is selected for each layer. The neural network parameters (i.e. the weight matrices and the biases) are found during training, typically using some variant of the stochastic gradient descent method. Note that we do not make use of any data pre-processing or post-processing as, instead, typical in a machine learning pipeline. Non dimensional units are used, though, for both the equinoctial elements and the mass).

\subsection{Architectures}
\label{sec:nnarchitecture}

The two fundamental architectures used in this paper are shown in Figure \ref{fig:ffnn}. The first one predicts the value function $v$, which in our case is the final optimal mass $m_f^*$, and hence is called \emph{value function network}, the second one predicts the throttle $u$ and the quantities $d_1, d_2, d_3$, which relate to the throttle direction via the relation $\hat {\mathbf i}_\tau = [d_r, d_t, d_n]^T / \vert[d_r, d_t, d_n]\vert$.
This second architecture is called \emph{policy network}. We found that the initialisation of the weights and biases is very important in the case studied and that bad initialisation leads to early convergence and poor performance in general both on the training and the validation sets. 
We used Kaiming Normal initialisation \cite{he2015delving} as this was found to be better suited to our network architectures. While such a normalization was originally developed for ReLu units, their functional similarity to the softplus units used here makes it appropriate for our architectures.

\subsection{Loss Functions}
\label{sec:learningapproaches}

We consider 6 different approaches to learning the state feedback optimal control. 
These can be classified into the two categories of \emph{Policy Imitation} and \emph{Value Function Approximation}.
Policy Imitation is straightforward in that the neural network learns the mapping between the spacecraft state and the optimal controls. On the other hand, Value Function Approximation is a more interesting strategy in that the neural network learns the mapping between the spacecraft state and the cost required to reach Venus' orbit: ideally, the neural network gradients with respect to the spacecraft state will then be make it possible to compute the optimal controls as well. Note that learning the value function, even when the following step of deducing the optimal controls fails, has its merit by itself as allows a number of applications, for example, in preliminary mission design where many transfer options need to be evaluated without going into the detail of the exact guidance law.

In order to describe the loss functions used for our networks, it is convenient to introduce the following components:
\begin{itemize}
    \item Policy Learning Loss Component - the error in the estimated controls: \\
    $l_{policy} = \mean{(u_\mathcal{N} - u^*)^2} + \mean{1 - \mathbf{\hat i}_{\mathcal{N}} \cdot \mathbf{\hat i}^*}$
    
    \item Value Function Loss Component - the error in the estimated value function: \\ 
    $l_{vf} = \mean{(J_{opt} - J_{\mathcal{N}})^2}$.
    
    \item Costate Loss Component - the error in the gradients of the network with respect to the true costates: \\ 
    $l_{\mathbf{\lambda}} = \mean{((\lambda_\mathbf{x})_{opt} - \nabla_\mathbf{x}J_{\mathcal{N}})^2}$.
    
    \item Hamiltonian Loss Component - the error in Hamiltonian computed from the network approximated costates: \\ 
    $l_{\mathcal{H}} = \mean{(\mathcal{H}(\mathbf{x}, m, \nabla_\mathbf{x}J_{\mathcal{N}}, \mathbf{u^*}))^2}$
    
    \item Control Loss Component - the error in the controls computed from the network approximated costates: \\ 
    $l_{\mathbf{u}} = \mean{(u(\mathbf{x}, m, \nabla_\mathbf{x}J_{\mathcal{N}}) - u^*)^2} + \mean{1 - \mathbf{\hat i}(\mathbf{x}, m, \nabla_\mathbf{x}J_{\mathcal{N}}) \cdot \mathbf{\hat i}^*}$
\end{itemize}
where we used the notation $\mean{\boldsymbol\cdot} = \frac{1}{N}\sum^N{(\boldsymbol\cdot)}$ to indicate a mean across the entire database $\mathcal D$. 
These loss components encapsulate a large variety of possible network training procedures which we here seek to study. The $l_{policy}$ component can be used to train a \emph{policy network} on the optimal policy whereas the remaining loss components can be used to train a \emph{value function network}. Combining the latter loss components in different ways we can conceive different training pipelines such as training the gradients of the \emph{value function network} directly on the $l_\mathbf{\lambda}$ component.




\begin{figure}[tb]
    \centering
    \includegraphics[width=0.85\columnwidth]{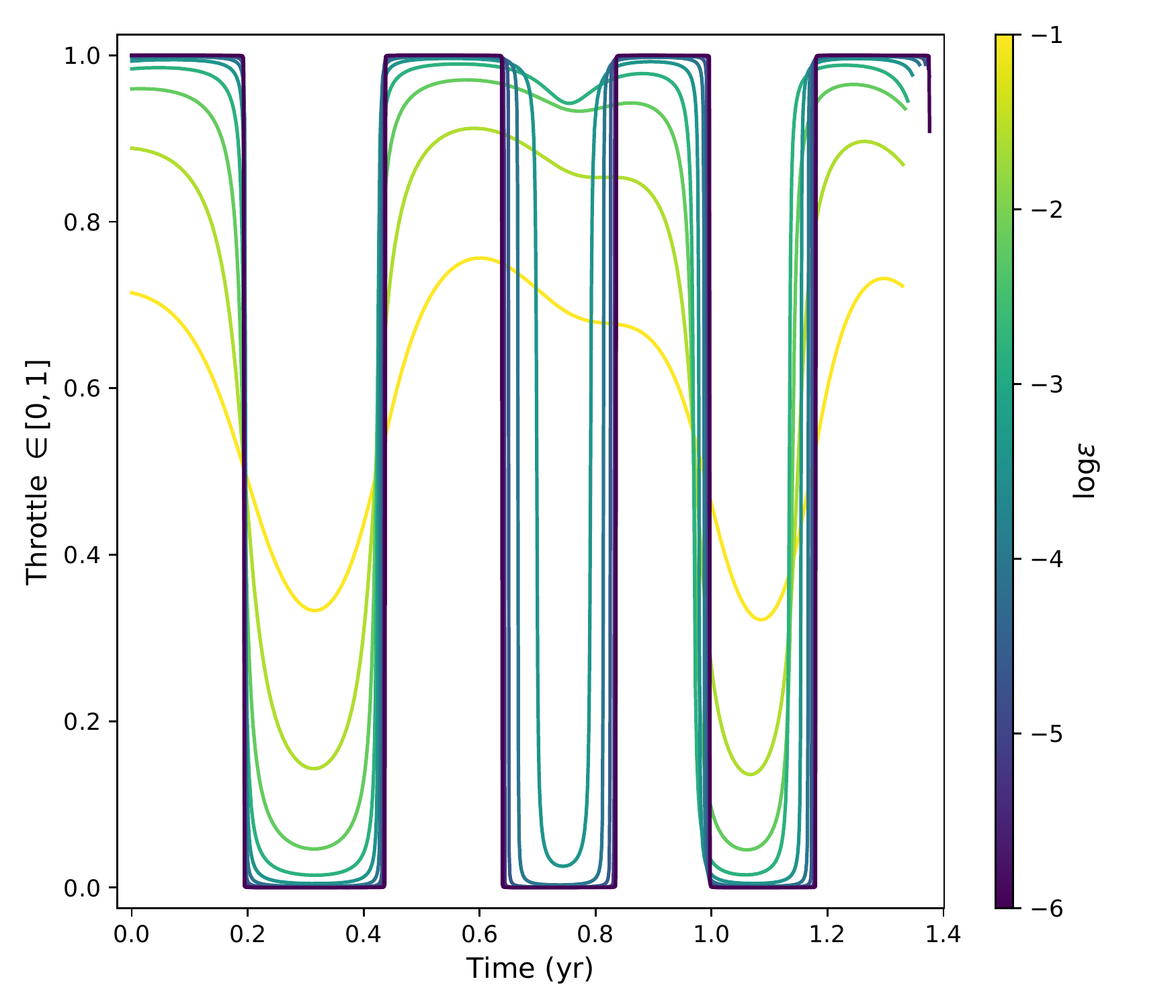}
    \caption{Solution of the two point boundary value problem for the throttle magnitude \emph{u(t)} and decreasing values of $\epsilon$.
    \label{fig:homotopy}}
\end{figure}

\section{Our training databases}
\label{sec:database_generation}
As discussed in Section \ref{sec:datageneration}, in order to build a database of trajectories, $\mathcal D$ using the \lq\lq Backward Generation of Optimal Examples\rq\rq  we require A) one nominal trajectory and B) a strategy to perturb the final augmented state. In the following two subsections we will be introducing these two items as defined in our experiments.

\subsection{The nominal trajectory}
\label{sec:nominaltrajectory}

To illustrate the concepts presented above and their potential we focus on one single interplanetary transfer, but it is to be remarked that the same methodology can be applied in general. We consider a spacecraft with mass $m_0 = \SI{1500}{\kilo\gram}$, a nuclear electric propulsion system specified by $I_{sp}=\SI{3800}{\second}$ and $c_1=\SI{0.33}{\newton}$. We compute the mass optimal transfer from the Earth to Venus orbit starting from the 7th of May 2005. Venus orbit is assumed keplerian and its orbital elements are computed at $\SI{1.05}{\year}$ from the launch date. The planet ephemerides are computed using JPL low-precision ephemerides \cite{standish2006keplerian}.

We solve the optimal control problem by solving the equation $\phi(\boldsymbol\lambda_0,\lambda_{m_0}, t_f)=0$ (see Eq.(\ref{eq:shooting})). Note that this is, essentially, a system of eight nonlinear equations in eight unknowns and can be solved by root finding methods (e.g. Powell, Levenberg-Marquardt) as well as by SQP or interior point methods (e.g. SNOPT \cite{snopt} or IPOPT \cite{ipopt}).

As it is well known (see \cite{haberkorn2004low} for example), the convergence radius for this problem can get rather small, to the point that if we were to try to directly solve the mass optimal problem (i.e. plugging $\epsilon=0$ in Eq.(\ref{eq:cost})) we would fail consistently as almost any initial guess on the co-state would not converge.

However, solving the problem for $\epsilon=0.1$ is reasonably simple as convergence is frequently achieved when starting with random co-states (we sample them from a uniform distribution with a standard deviation of 10). Note that we use nondimensional units for the state so that the astronomical unit AU is used for length, the spacecraft initial mass for mass, and the rest is set as to get $\mu=1.$.

Gradually decreasing $\epsilon$ from $0.1$ down to $10^{-6}$ (as visualised in Figure~\ref{fig:homotopy}), allows us to obtain the final mass optimal trajectory which is visualised in Figure~\ref{fig:nominal_flights:nominal_trajectory}. We refer to the final trajectory (with $\epsilon=10^{-6}$) as the \emph{nominal trajectory} in the following.

The nominal trajectory reaches the orbit of Venus after $t^*_f=\SI{1.376}{\year}$ and spends $m_p=\SI{210.47}{\kilo\gram}$ of propellant and is visualized in Figure \ref{fig:nominal_flights}.

\subsection{Perturbation and Database Size}
\label{sec:dbvariation}

It takes in the order of minutes on an Intel Xeon E5-2650L v4 processor at $\SI{1.70}{\giga\hertz}$ to solve (completing the whole homotopy path) one optimal control problem. And this is assuming the initial guess on the co-states are withing the radius of convergence of the shooting function solver. The largest database we trained on consists of about ${10}^{6}$ trajectories, thus brute-forcing the database generation would require on the order of years without multi-threading. Using, instead, the \lq\lq Backward Generation of Optimal Examples\rq\rq (see Section \ref{sec:computationcost}) a database containing ${10}^{6}$ optimal trajectories is generated in $\sim\SI{6}{\hour}$: an improvement of several orders of magnitude.

Overall, we generated a total of 7 databases around the same nominal trajectory by varying the perturbation sizes (of the final augmented state) and number of trajectories. Table~\ref{table:databases} shows the databases we generated and the three parameters that characterise each database: the perturbation size, the number of sample points along a trajectory and the number of trajectories generated.

\begin{table}[tb]
\sisetup{group-separator={,},
         output-decimal-marker={.},
         group-four-digits,
         per-mode=fraction,
         group-digits=integer,
         round-mode=places,
         round-precision=1}
    \begin{center}
    \renewcommand{\arraystretch}{0.8}
    \setlength{\tabcolsep}{2.0pt}
         \begin{tabular}{@{}lSSSSSSS@{}}
         \toprule
         Database Name                     & \multicolumn{1}{c}{\emph{A}} & \multicolumn{1}{c}{\emph{B}} & \multicolumn{1}{c}{\emph{C}} & \multicolumn{1}{c}{\emph{D}} & \multicolumn{1}{c}{\emph{E}} & \multicolumn{1}{c}{\emph{F}} & \multicolumn{1}{c}{\emph{G}} \\
         \cmidrule{2-8}
         $\rho$                           & \num{0.2} & \multicolumn{2}{c}{\num{0.4}} & \num{5} & \num{20} & \multicolumn{2}{c}{custom} \\
         \cmidrule{2-8}
         \# of Samples                    & \multicolumn{1}{c}{\num{100}} & \multicolumn{1}{c}{\num{100}} & \multicolumn{1}{c}{\num{100}} & \multicolumn{1}{c}{\num{128}}  & \multicolumn{1}{c}{\num{100}} & \multicolumn{1}{c}{\num{100}} & \multicolumn{1}{c}{\num{100}} \\
         \# of Trajectories Generated     & \multicolumn{1}{c}{\num{500000}} & \multicolumn{1}{c}{\num{500000}} & \multicolumn{1}{c}{\num{1000000}} & \multicolumn{1}{c}{\num{400000}}  & \multicolumn{1}{c}{\num{1000000}} & \multicolumn{1}{c}{\num{1000000}} & \multicolumn{1}{c}{\num{3588120}} \\
         \# of Trajectories Succeeded     & \multicolumn{1}{c}{\num{429316}} & \multicolumn{1}{c}{\num{382193}} & \multicolumn{1}{c}{\num{764479}} & \multicolumn{1}{c}{\num{265603}}  & \multicolumn{1}{c}{\num{409076}} & \multicolumn{1}{c}{\num{557395}} & \multicolumn{1}{c}{\num{999985}} \\
         \cmidrule{2-8}
         Total Size of Database & \multicolumn{1}{c}{\num{42931600}} & \multicolumn{1}{c}{\num{38219300}} & \multicolumn{1}{c}{\num{76447900}} & \multicolumn{1}{c}{\num{33997184}}  & \multicolumn{1}{c}{\num{40907600}} & \multicolumn{1}{c}{\num{55739500}} & \multicolumn{1}{c}{\num{99998500}} \\ \bottomrule
        \end{tabular}
    \end{center}
    \caption{Parameters of the different databases.}
    \label{table:databases}
\end{table}

For all databases except for \emph{F} and \emph{G}, we used a fixed perturbation size for all the parameters. For databases \emph{F} and \emph{G} we experimented with a non uniform perturbation size tailored to create a visually dense database around the conditions of interest. The exact perturbations of each component are listed in Table~\ref{table:perturbation_sizes}. Additionally the trajectories in \emph{F} and \emph{G} were terminated whenever the spacecraft semimajor axis went outside $\left[a_{Venus} - 100\times r_{Venus}, a_{Earth} + 100\times r_{Earth}\right]$ and the inclination went outside of $\left[-7^{\circ},7^{\circ}\right]$. This was done to avoid including trajectory segments in the database that would confuse the training and be well outside the region of interest. (The region of interest can be loosely defined as the torus encompassing Earth and Venus' orbits.)
.
\begin{table}[tb]
    \begin{center}
    \renewcommand{\arraystretch}{0.8}
    \setlength{\tabcolsep}{5pt}
         \begin{tabular}{@{}lSSSSSS@{}}\toprule
          & m & $\lambda_p$ & $\lambda_f$ & $\lambda_g$ & $\lambda_h$ & $\lambda_k$ \\
         \cmidrule{2-7}
         mean               & 0.0  & 0.0 & 0.0 & 0.0 & 0.0 & 0.0 \\
         standard deviation & 0.01 & 5.0 & 1.0 & 1.0 & 0.0 & 0.0 \\ \bottomrule
        \end{tabular}
    \end{center}
    \caption{Perturbations of each component in the creation of databases \emph{F} and \emph{G}.}
    \label{table:perturbation_sizes}
\end{table}

\begin{figure}[tb]
    \centering
    \subfloat[Above Plane View]{\includegraphics[width=0.999\textwidth]{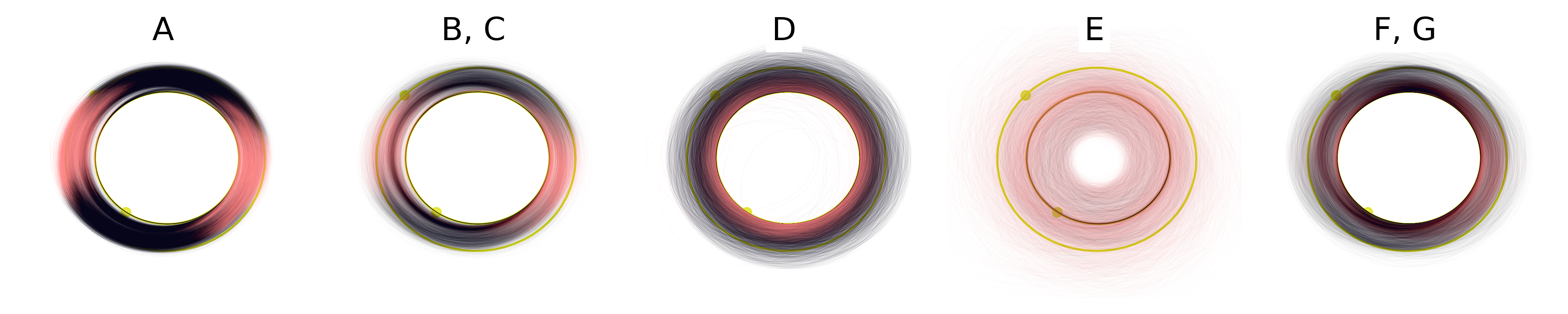}} \\
    \subfloat[In-Plane View]{\includegraphics[width=0.999\textwidth]{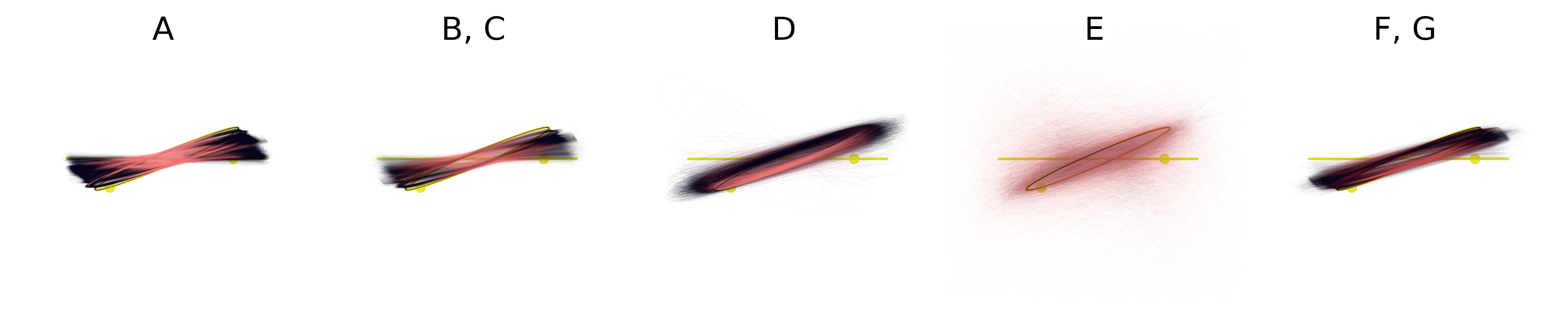}}
    \caption{Visualization of the trajectories in the generated databases. Thrust arcs are indicated in pink. The Earth and Venus orbits are visualized to provide some sense of scale.
    \label{fig:overlapping_databases}}
\end{figure}

Figure~\ref{fig:overlapping_databases} shows a visual representation of each of the generated databases from the above plane and in-plane views.

\section{Network Training}
\label{sec:training}

We experimented with several combinations of loss functions and architectures and we report here the results on 4 different networks which we found particularly significant. 
The first trained network, indicated $\mathcal{N}_1$ is a \emph{policy network} with 3 hidden layers and 200 neurons per layer. 
The following  $\mathcal{N}_2$ and $\mathcal{N}_3$ are \emph{value function network}s with 9 hidden layers and 200 neurons per layer. 
A final network, $\mathcal{N}_4$, is also a \emph{value function network}, but this time with 3 hidden layers and 1000 neurons per layer.
The loss function used to train the various networks is constructed out of the components defined in Section \ref{sec:learningapproaches} as detailed in Eq.~\eqref{eq:loss_functions}.

\begin{subequations}
    \label{eq:loss_functions}
    \begin{align}
    \label{eq:loss_functions:policy}
    l_{\mathcal{N}_1} &= l_{policy}                    \\
    \label{eq:loss_functions:value_function_only}
    l_{\mathcal{N}_2} &= l_{vf}                        \\
    \label{eq:loss_functions:value_function_and_gradients}
    l_{\mathcal{N}_3} &= l_{vf} + l_{\mathbf{\lambda}} \\
    \label{eq:loss_functions:value_function_and_hamiltonian_ctrl}
    l_{\mathcal{N}_4} &= l_{vf} + s_1 l_{\mathcal{H}} + l_{\mathbf{u}}
    \end{align}
\end{subequations}
where $s_1$ is a scaling parameter chosen to normalise the loss components relative to each other. We find that the $l_\mathcal{H}$ component is generally poorly scaled relative to the other components. Depending on the network initialisation, we found that, for our particular architecture and initialisation, the $l_\mathcal{H}$ component started with a value of $\approx {10}^{-1}$ which was too small relative to the other components (which were of the order $10^0-10^1$). For this reason we selected a value of $s_1=10^2$. 

These four networks were trained using the same optimiser with similar hyper-parameters. Specifically, we used the Amsgrad \cite{reddi2018adamconv} optimiser with $\beta_1=0.9$, $\beta_2=0.999$, $\epsilon={10}^{-8}$ and $weight\ decay=0.0$. The networks $\mathcal{N}_1$, $\mathcal{N}_2$ and $\mathcal{N}_3$ were trained with $lr={10}^{-4}$ whereas $\mathcal{N}_4$ was trained with $lr={10}^{-3}$.

Based on the improvements in the validation loss, we fixed the number of training epochs for each network type for all databases. $\mathcal{N}_1$ was trained for $250$ epochs, $\mathcal{N}_2$ and $\mathcal{N}_4$ were trained for $100$ epochs, and $\mathcal{N}_3$ was trained for $1000$ epochs. The minibatch size was also fixed at $4096$ examples per GPU and multi-GPU training with up to 4 GPUs was utilised where possible.

We split the databases into training, validation and test sets following an 80 - 10 - 10 split. The nominal trajectory was excluded from the training as we wanted to measure whether the networks generalised to the underlying, unseen reference trajectory. Furthermore, using the validation split, we incorporated the reduction of the learning rate whenever the validation loss plateaued.

\begin{figure}[tb]
    \centering
    \subfloat[Nominal trajectory to Venus orbit]{
        \centering
        \includegraphics[width=0.3\columnwidth]{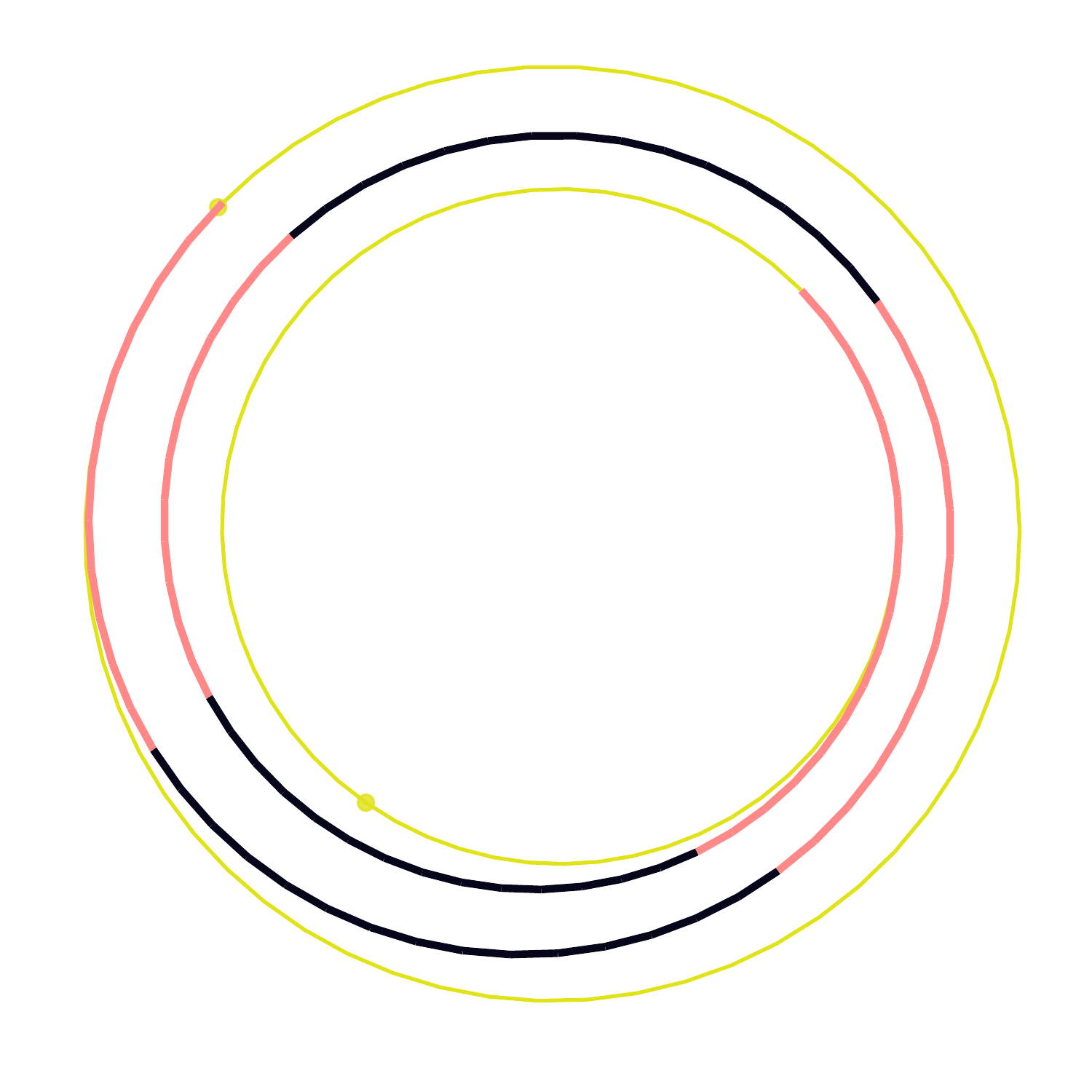}
        \label{fig:nominal_flights:nominal_trajectory}
    }
    \subfloat[Policy Imitation]{
        \centering
        \includegraphics[width=0.3\columnwidth]{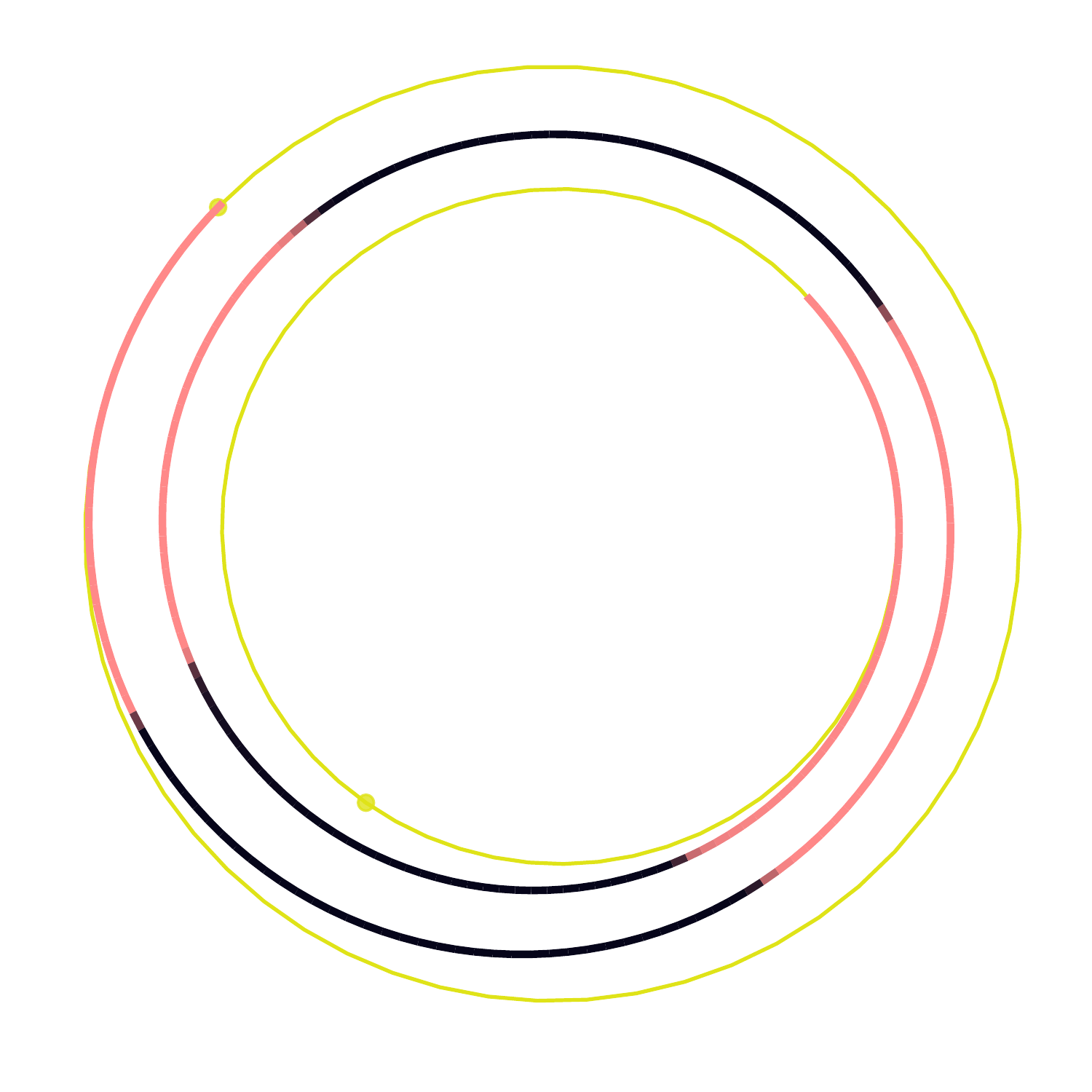}
        \label{fig:nominal_flights:optimal_policy}
    }
    \subfloat[Value Function without Gradients]{
        \centering
        \includegraphics[width=0.3\columnwidth]{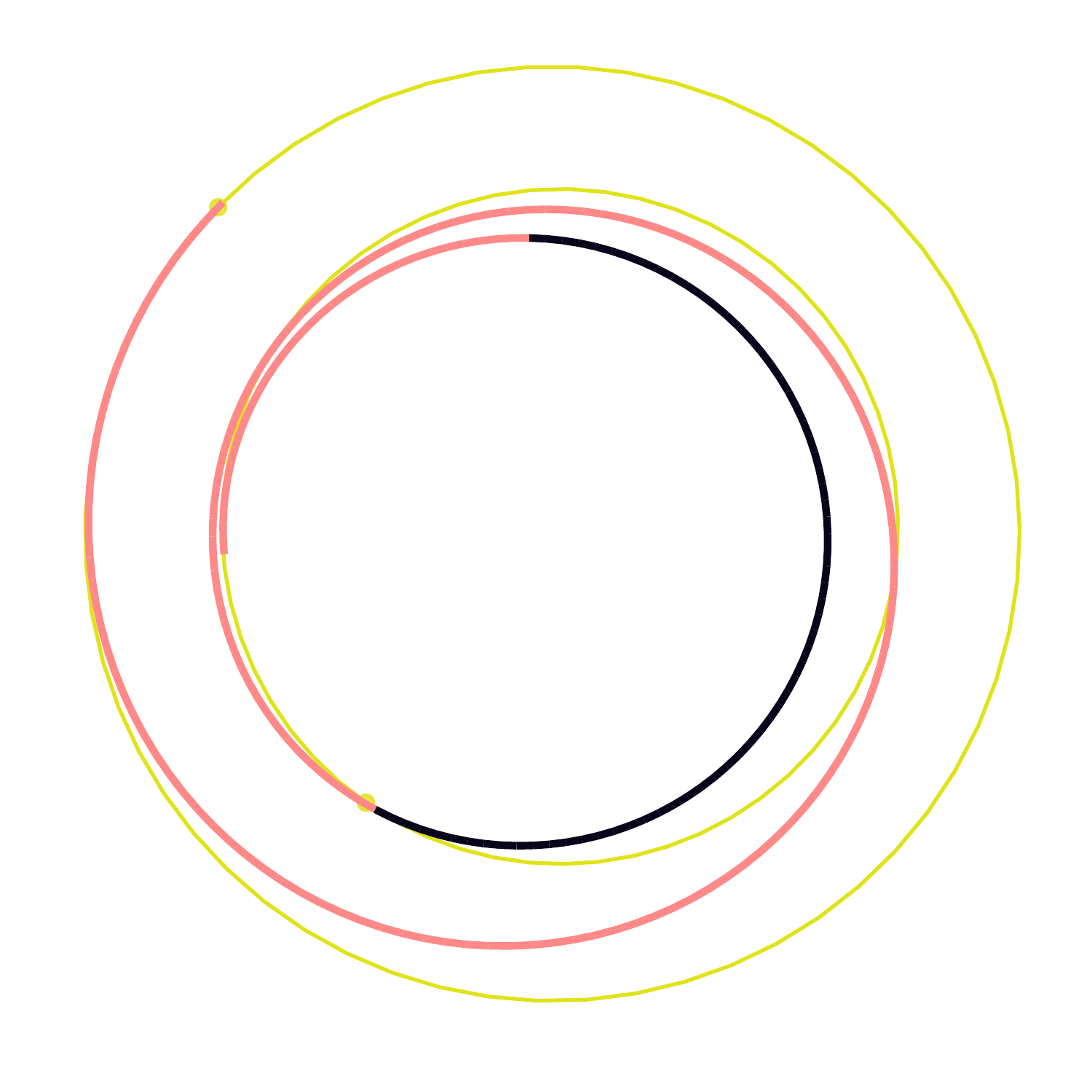}
        \label{fig:nominal_flights:vf_no_grad}
    }
    \newline
    \subfloat[Value Function with Gradients]{
        \centering
        \includegraphics[width=0.3\columnwidth]{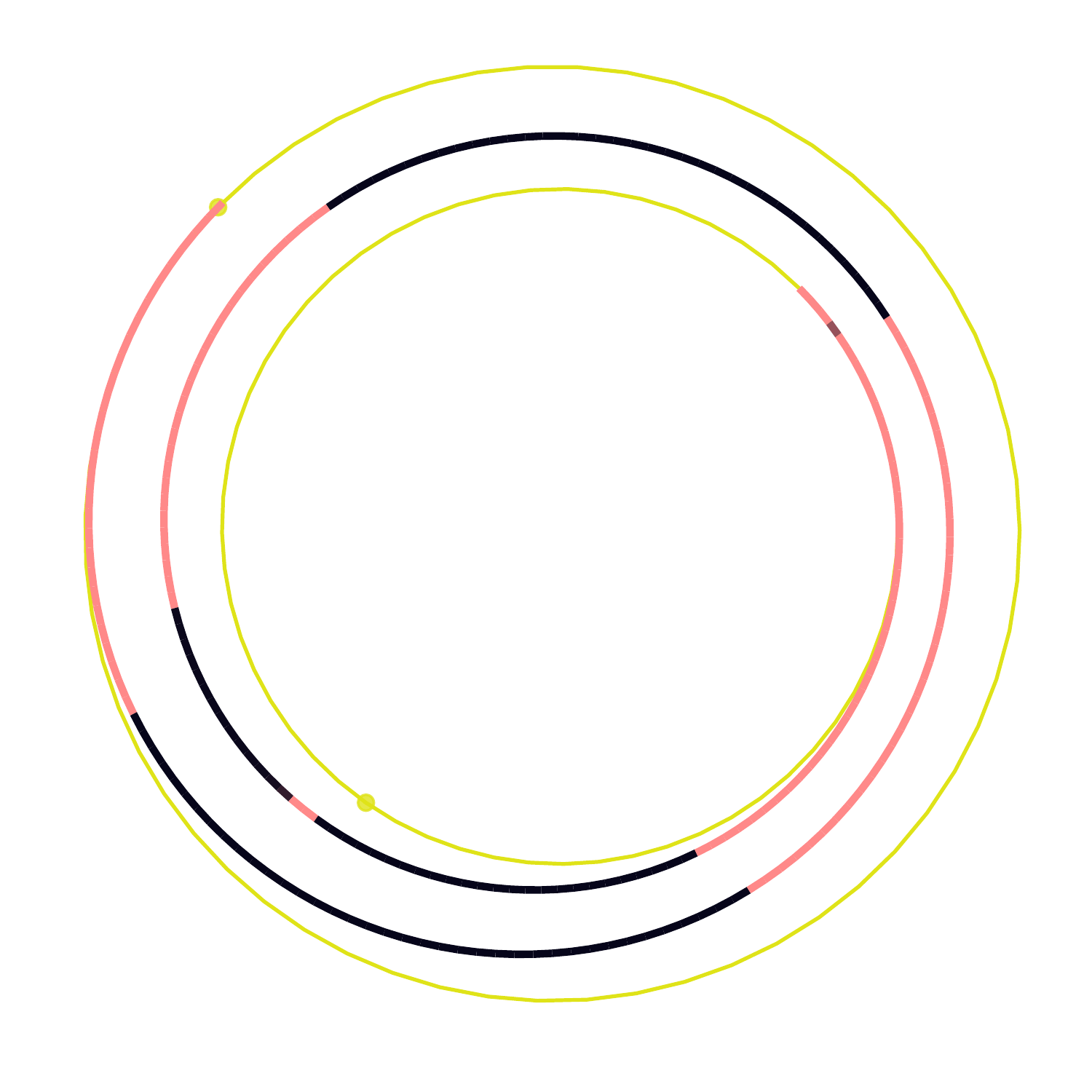}
        \label{fig:nominal_flights:vf_w_grad}
    }
    \subfloat[Value Function with Hamiltonian and Controls]{
        \centering
        \includegraphics[width=0.3\columnwidth]{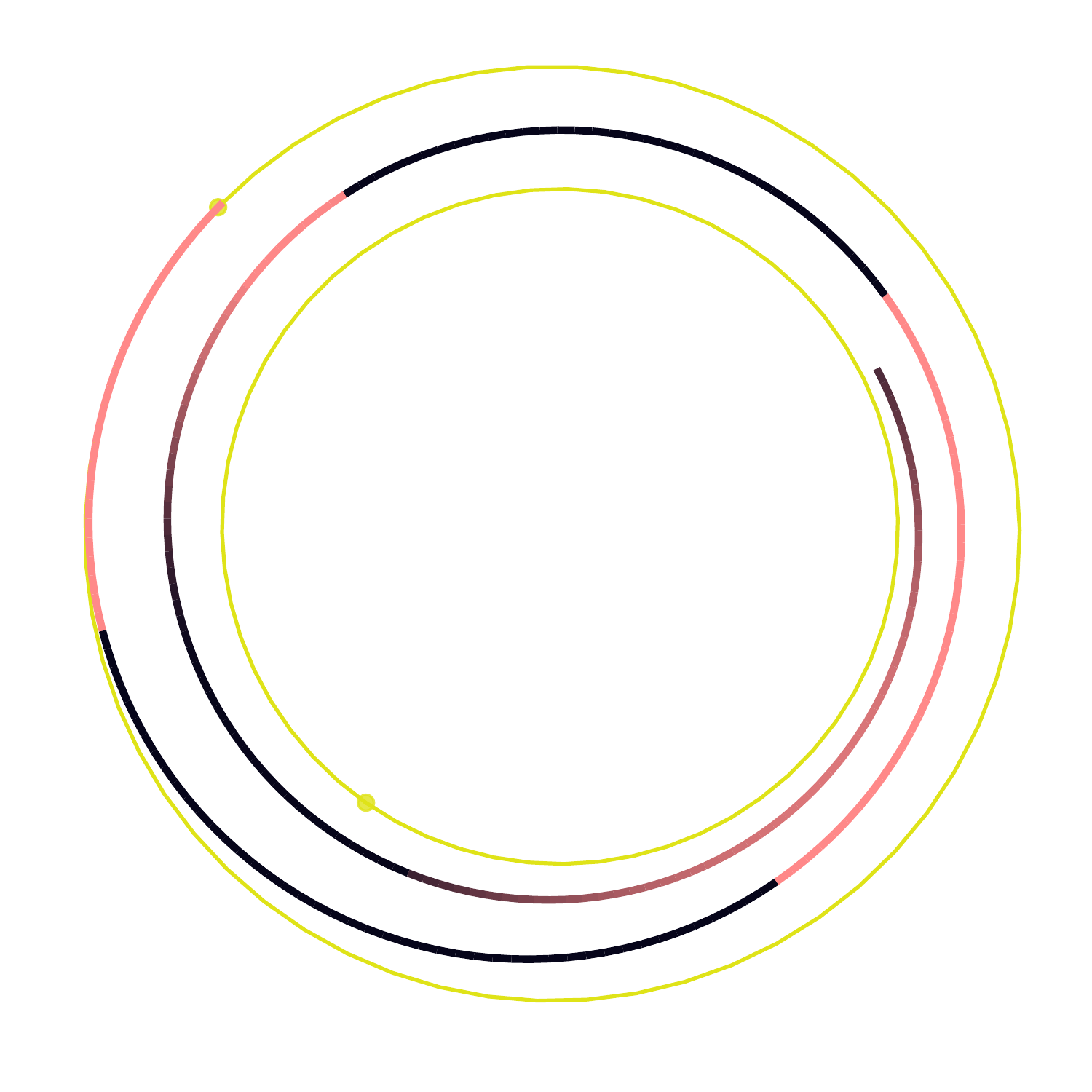}
        \label{fig:nominal_flights:vf_w_hamil_ctrl}
    }
    \caption{Nominal trajectories generated by G\&CNETs trained on \emph{G}. Thrust arcs are indicated in a lighter colour.\label{fig:nominal_flights}}
\end{figure}

\section{Results}
\label{sec:results}
There are numerous ways to evaluate the performance of the trained G\&CNETs, each with their pros and cons. Here we look at two main elements: A) the G\&CNET performance when controlling the spacecraft starting from the nominal trajectory initial conditions, B) the G\&CNET performance when controlling the spacecraft starting from \lq\lq any\rq\rq\ initial conditions. 


\subsection{Performance along the nominal conditions}
\label{sec:nominalperf}
The first performance criteria is to look at how closely and optimally the spacecraft reaches the orbit of Venus when starting from the nominal conditions used to generated the different databases via our\lq\lq backward generation of optimal examples\rq\rq\ technique. In Figure~\ref{fig:nominal_flights} the interplanetary transfer resulting from using the networks is shown in comparison to the nominal transfer (i.e. the optimal transfer) revealing different levels of performances.
In order to quantify the comparison, we first look at the final Euclidean distance between the first 5 equinoctial elements ($p,f,g,h,k$) to the target equinoctial elements defining Venus orbit. We refer to this measure as the \lq\lq reduced Euclidean distance\rq\rq\ (rEd) and denote it with $\norm{\Delta(\mathbf x)}{2}$. 
The rEd is measured between Venus orbit and the orbit achieved after numerically integrating for the optimal time $t^*_f$ Eq.(\ref{eq:eom}) where $f_r, f_t, f_n$ are computed from the G\&CNET. Table~\ref{table:nominal_performance_closeness} shows the rEd measure associated to each network controller trained on each of the databases. For scale, the rEd measure between the Earth and Venus orbit is $\num{0.28}$.

\begin{table}[tb]
\sisetup{group-separator={,},
         output-decimal-marker={.},
         group-four-digits,
         per-mode=fraction,
         group-digits=integer,
         round-mode=figures,
         round-precision=2}
    \begin{center}
    \renewcommand{\arraystretch}{0.8}
    \setlength{\tabcolsep}{0.5pt}
         \begin{tabular}{@{}clSSSSSSS@{}}
         \toprule
         & Training Database           & \multicolumn{1}{c}{\emph{A}} & \multicolumn{1}{c}{\emph{B}} & \multicolumn{1}{c}{\emph{C}} & \multicolumn{1}{c}{\emph{D}} & \multicolumn{1}{c}{\emph{E}} & \multicolumn{1}{c}{\emph{F}} & \multicolumn{1}{c}{\emph{G}} \\
         \cmidrule{2-9}
         \parbox[t]{3mm}{\multirow{4}{*}{\rotatebox[origin=c]{90}{NN Arch.}}}
         & \emph{policy network} {} {} {} {} {} {} {} {} {} {} {} {} {} {} $\mathcal{N}_1$ & 0.0008496957382066033 & 0.0009365485605023099 & 0.0003554583573651388 & 0.00789883582692545    & 0.11771242654694547    & 0.0035125428849142135 & 0.004079742498031989  \\
         & \emph{value function network} {} $\mathcal{N}_2$   & 0.22959604380299325   & 0.23466345374491782   & 0.22917243562648903   & 0.10875723732005131    & 0.0788302706171756     & 0.21585004624378187   & 0.2080345929251668    \\
         & \emph{value function network} {} $\mathcal{N}_3$   & 0.006166496425087346  & 0.010799662005421352  & 0.006281218491072838  & 0.0015207812747025972  & 0.00045372158996108706 & 0.006971715237202418  & 0.0012619602444991428 \\
         & \emph{value function network} {} $\mathcal{N}_4$   & 0.005893410049048961  & 0.017259112660384126  & 0.017143280807155988  & 0.061819055598567936   & 0.09877101020277218    & 0.04728013359819654   & 0.04117914332132499   \\ \bottomrule
        \end{tabular}
    \end{center}
    \caption{Reduced Euclidean distance (rEd) for the trained networks across databases.}
    \label{table:nominal_performance_closeness}
\end{table}

Note that the rEd distance is not, alone, returning the full picture on the network performances since it does not include any information on the propellant used.
In order to evaluate the mass optimality resulting from the various networks, we solve the optimal control problem described in Section \ref{sec:tpbvp} from the spacecraft state at $t^*_f$ to Venus' orbit at $t^*_f+\Delta t$, and compare it to the fixed time optimal transfer from Earth to Venus, the fixed time being $t^*_f+\Delta t$. The resulting difference in propellant used is called propellant discrepancy and is the indicator we use to quantify the mass optimality of a network controller. Table~\ref{table:nominal_performance_completion} shows the propellant discrepancy of each controller trained on each database.

%
%
%
%
%
%
%
%

\begin{table}[tb]
\sisetup{group-separator={,},
         output-decimal-marker={.},
         group-four-digits,
         per-mode=fraction,
         group-digits=integer,
         round-mode=places,
         round-precision=2}
    \begin{center}
    \renewcommand{\arraystretch}{0.8}
    \setlength{\tabcolsep}{4.0pt}
         \begin{tabular}{@{}clSSSSSSS@{}}
         \toprule
         & Training Database           & \multicolumn{1}{c}{\emph{A}} & \multicolumn{1}{c}{\emph{B}} & \multicolumn{1}{c}{\emph{C}} & \multicolumn{1}{c}{\emph{D}} & \multicolumn{1}{c}{\emph{E}} & \multicolumn{1}{c}{\emph{F}} & \multicolumn{1}{c}{\emph{G}} \\
         \cmidrule{2-9}
         \parbox[t]{3mm}{\multirow{4}{*}{\rotatebox[origin=c]{90}{NN Arch.}}}
         & \emph{policy network} {} {} {} {} {} {} {} {} {} {} {} {} {} {} $\mathcal{N}_1$ & 0.43798841160819224  & 1.5236959595195665  & 0.8013400643753776  & 20.731903683839313   & 6.549796967430355  & 1.4036122486866809  & 1.2216856254240738 \\
         & \emph{value function network} {} $\mathcal{N}_2$   & 234.52587285683163   & 279.0425917818828   & 268.9574449511013   & 2.560973782317233    & 24.32043573811471  & 216.99269303652963  & 227.26413653233203 \\
         & \emph{value function network} {} $\mathcal{N}_3$   & 5.975085445177275    & 8.258895561228385   & 6.242073215053045   & 9.976365206889437    & 14.739134037308588 & 7.167784591070037   & 1.1921199353462364 \\
         & \emph{value function network} {} $\mathcal{N}_4$   & 7.393927459839134    & 11.583229808548868  & 13.141434941176367  & 4.2335097789213965   & 6.915415992329088  & 8.541549419049554    & 6.5084866960102294 \\ 
         \bottomrule
        \end{tabular}
    \end{center}
    \caption{Propellant discrepancy for various networks and across databases. For scale the nominal optimal propellant is 210.47 kg.}
    \label{table:nominal_performance_completion}
\end{table}

{
}

\subsection{Performance away from nominal conditions}
\label{sec:testperf}
A second performance criteria is to look at the behaviour of the network controllers when initial conditions are considered that are far away from the nominal trajectory used to generated the databases via our\lq\lq backward generation of optimal examples\rq\rq\ technique. For this criteria we consider two indicators we think cover the most important characteristics of a controller: first we look at mean errors of the optimal policy predictions as computed from a G\&CNET, and then we look at how close to Venus' orbit each spacecraft eventually gets when perturbing the initial nominal conditions by increasing factors. Additionally, for the \emph{value function networks}, we evaluate the ability of the networks to predict the optimal value function (i.e. to compute the optimal propellant to reach Venus orbit from any spacecraft state).

Table \ref{table:test_set_performance} shows the mean errors of the controls computed from each neural network on their test sets. 
We chose to use the mean absolute error for the throttle and the mean angular error for the thrust direction. 
The mean angular error is useful as it foregoes issues of vector scaling and focuses on the important aspect of the thrust vector, namely, the direction, whereas the throttle error focuses on the throttle magnitude.
The throttle error is thus defined as $\Delta u = \lvert u_\mathcal{N} - u^*\rvert$ and the angular error is defined as $\psi_{i_{\tau}} = \arccos{\left[\mathbf{\hat i_{\tau\mathcal{N}}} \cdot \mathbf{\hat i_\tau^*}\right]}$. 
We denote the mean using the notation $\mean{\boldsymbol\cdot} = \frac{1}{N}\sum^N{(\boldsymbol\cdot)}$ and the standard deviation using the notation $\std{\boldsymbol\cdot} = \frac{1}{N-1}\sqrt{\sum^N{(\boldsymbol\cdot - \mean{\boldsymbol\cdot})}}$, e.g. $\mean{\Delta u}$ is the mean throttle error and $\std{\Delta u}$ is the standard deviation of the throttle error.
We will use this notation in the rest of this paper.
In Table \ref{table:test_set_performance} we see that the \emph{policy network} $\mathcal{N}_1$ and \emph{value function network} $\mathcal{N}_3$ have consistently low errors with some variability across databases and a few exceptions.
We also note that the \emph{value function network} $\mathcal{N}_2$ has a high error in general. This network is trained using a loss function $l_{\mathcal{N}_2}$ that does not penalise errors in the value function gradients, this results in a network that is unusable for the purpose of reconstructing the optimal policy. 
By adding such a contribution to the loss function, we get $l_{\mathcal{N}_3}$ which has, instead, a low prediction error for the optimal policy. 
To a lesser extent, we see observe the same improvement in $\mathcal{N}_4$ which uses a loss $l_{\mathcal{N}_4}$ that while not enforcing the value function gradient directly, it does penalize violations to the Belmann equation Eq.(\ref{eq:HJB2}) and the transversality condition on the Hamiltonian (free time transfer). 
The advantage of such a loss is that it does not require the co-states and can in principle be applied to learn from optimal examples generated by direct methods too.

\begin{table}[tb]
\sisetup{group-separator={,},
         output-decimal-marker={.},
         group-four-digits,
         per-mode=fraction,
         group-digits=integer,
         round-mode=figures,
         round-precision=2,
         separate-uncertainty}
    \begin{center}
    \renewcommand{\arraystretch}{0.8}
    \setlength{\tabcolsep}{3.0pt}
         \begin{tabular}{@{}clcSSSSSSS@{}}
         \toprule
         & Training Database           &  & \multicolumn{1}{c}{\emph{A}} & \multicolumn{1}{c}{\emph{B}} & \multicolumn{1}{c}{\emph{C}} & \multicolumn{1}{c}{\emph{D}} & \multicolumn{1}{c}{\emph{E}} & \multicolumn{1}{c}{\emph{F}} & \multicolumn{1}{c}{\emph{G}} \\
         \cmidrule{2-10}
         
         \parbox[t]{3mm}{\multirow{8}{*}{\rotatebox[origin=c]{90}{NN Architectures}}}
         & \emph{policy network} {} {} {} {} {} {} {} {} {} {} {} {} {} {} $\mathcal{N}_1$ & $\mean{\Delta u}$                  
         & 0.02747097492113172       & 0.03749736709317607       & 0.04371366886214349       & 0.06185868828897471       & 0.12366451047878498        & 0.04247776942890010         & 0.035121 \\
         & & $\mean{\psi_{i_\tau}}$
         & \ang{0.52264117088310724} & \ang{0.44102619771202484} & \ang{0.67714353917903936} & \ang{7.23199879388999989} & \ang{10.78540202196060172} & \ang{1.41550592235982986}   & \ang{1.351423} \\
         
         & \emph{value function network} {} $\mathcal{N}_2$   & $\mean{\Delta u}$
         & 0.45899101989126151       & 0.39011120630699453       & 0.41972713517605875        & 0.18788658475710182       & 0.12563704453506211        & 0.43001550879897527        & 0.416265 \\
         & & $\mean{\psi_{i_\tau}}$
         & \ang{11.17524754460517578} & \ang{8.76020678820964704} & \ang{9.18732920436990064} & \ang{16.44489423802326655} & \ang{14.93199376349494933} & \ang{8.73538658529327350}  & \ang{8.631974} \\
         
         & \emph{value function network} {} $\mathcal{N}_3$   & $\mean{\Delta u}$
         & 0.04111632433196261       & 0.05675582096733024       & 0.04940776699639186        & 0.02888650513635980 & 0.02920292068872990         & 0.04586018373574700        & 0.030071 \\
         & & $\mean{\psi_{i_\tau}}$
         & \ang{0.26375799993785409} & \ang{0.52526804493996171} & \ang{0.49044289465467744}  & \ang{3.70662731413318802} & \ang{4.88159242343009403}   & \ang{0.91724847433761914}  & \ang{0.454217} \\
                              
         & \emph{value function network} {} $\mathcal{N}_4$   & $\mean{\Delta u}$
         & 0.06801590074729393       & 0.10553454820472466       & 0.09600741300635994       & 0.07775136880926968       & 0.20973984474414484        & 0.11909338425456234         & 0.107057 \\
         & & $\mean{\psi_{i_\tau}}$ 
         & \ang{2.74645505552110913} & \ang{2.95352592139228287} & \ang{2.74425219833024814} & \ang{19.04698327090601850} & \ang{17.10704315471388526} & \ang{3.95213078080952629}   & \ang{4.495703} \\
         \bottomrule
        \end{tabular}
    \end{center}
    \caption{Mean absolute error of the controls computed by the controllers on the test set of their respective training databases}
    \label{table:test_set_performance}
\end{table}

Table \ref{table:test_set_performance_value_function} shows the performance of the \emph{value function networks} on predicting the optimal value function for the initial conditions in the test set of their training databases. We show the mean absolute error in terms of the propellant required to reach Venus' orbit optimally from the given initial conditions, and we denote this by $\mean{\Delta J} = \mean{\lvert J_\mathcal{N} - J^*\rvert}$.
In this case we see, unsurprisingly, that the \emph{value function network} $\mathcal{N}_2$, using a loss function that only cares about the value function value, outperforms the others with an accuracy in predicting the propellant required with an accuracy on the order of \SI{2}{\kilo\gram} for the case of a training on the database G.

\begin{table}[tb]
\sisetup{group-separator={,},
         output-decimal-marker={.},
         group-four-digits,
         per-mode=fraction,
         group-digits=integer,
         round-mode=figures,
         round-precision=3,
         separate-uncertainty}
    \begin{center}
    \renewcommand{\arraystretch}{0.8}
    \setlength{\tabcolsep}{2.5pt}
         \begin{tabular}{@{}clcSSSSSSS@{}}
         \toprule
         & Training Database           &  & \multicolumn{1}{c}{\emph{A}} & \multicolumn{1}{c}{\emph{B}} & \multicolumn{1}{c}{\emph{C}} & \multicolumn{1}{c}{\emph{D}} & \multicolumn{1}{c}{\emph{E}} & \multicolumn{1}{c}{\emph{F}} & \multicolumn{1}{c}{\emph{G}} \\
         \cmidrule{2-10}
         
         \parbox[t]{3mm}{\multirow{6}{*}{\rotatebox[origin=c]{90}{NN Architectures}}}
         & \emph{value function network} {} $\mathcal{N}_2$   & $\mean{\Delta J}$
         & 3.6147734687004154 & 3.99716042209992 & 4.467361554200806 & 21.021967343566544 & 131.32670787369204 & 2.305596722911365 & 1.98070610970204 \\
         & & $\std{\Delta J}$
         & 4.1153613603293095 & 4.24626472195563 & 4.41633933336513 & 36.43151466410789 & 185.28783393764368 & 2.487279865151205 & 2.253415553696085 \\
         
         & \emph{value function network} {} $\mathcal{N}_3$   & $\mean{\Delta J}$
         & 4.77030583673943 & 8.1932703669378    & 7.45500030075969 & 42.51114462940893 & 165.95365088248587 & 13.88629416492105 & 8.95964890710714 \\
         & & $\std{\Delta J}$
         & 8.06621419058301 & 12.233079593651265 & 11.444301453744421 & 88.1582179999088 & 331.99244501057433 & 19.753833153946783 & 15.191844305887498 \\

         & \emph{value function network} {} $\mathcal{N}_4$   & $\mean{\Delta J}$
         & 77.07726396777082 & 88.69858992347808 & 69.59784611812425 & 80.5800568797437 & 392.2784420586457  & 109.76701130918805 & 97.04308576049343 \\
         & & $\std{\Delta J}$
         & 59.28804701382764 & 68.94572003799725 & 54.694185554022255 & 147.81246054809847 & 382.8482998863169  & 85.19564367500116 & 75.65934401662157 \\
        \bottomrule
        \end{tabular}
    \end{center}
    \caption{Mean absolute error of the value function (final propellant mass in kilograms) computed by the value function networks on the test set of their respective training datasets.}
    \label{table:test_set_performance_value_function}
\end{table}

\begin{figure}[tb]
    \centering
    \subfloat[$\mathcal A_2$: perturbation of 2\%\label{fig:final_orbit_halos:vf_w_grad:2p}]{
        \centering
        \includegraphics[width=0.80\columnwidth]{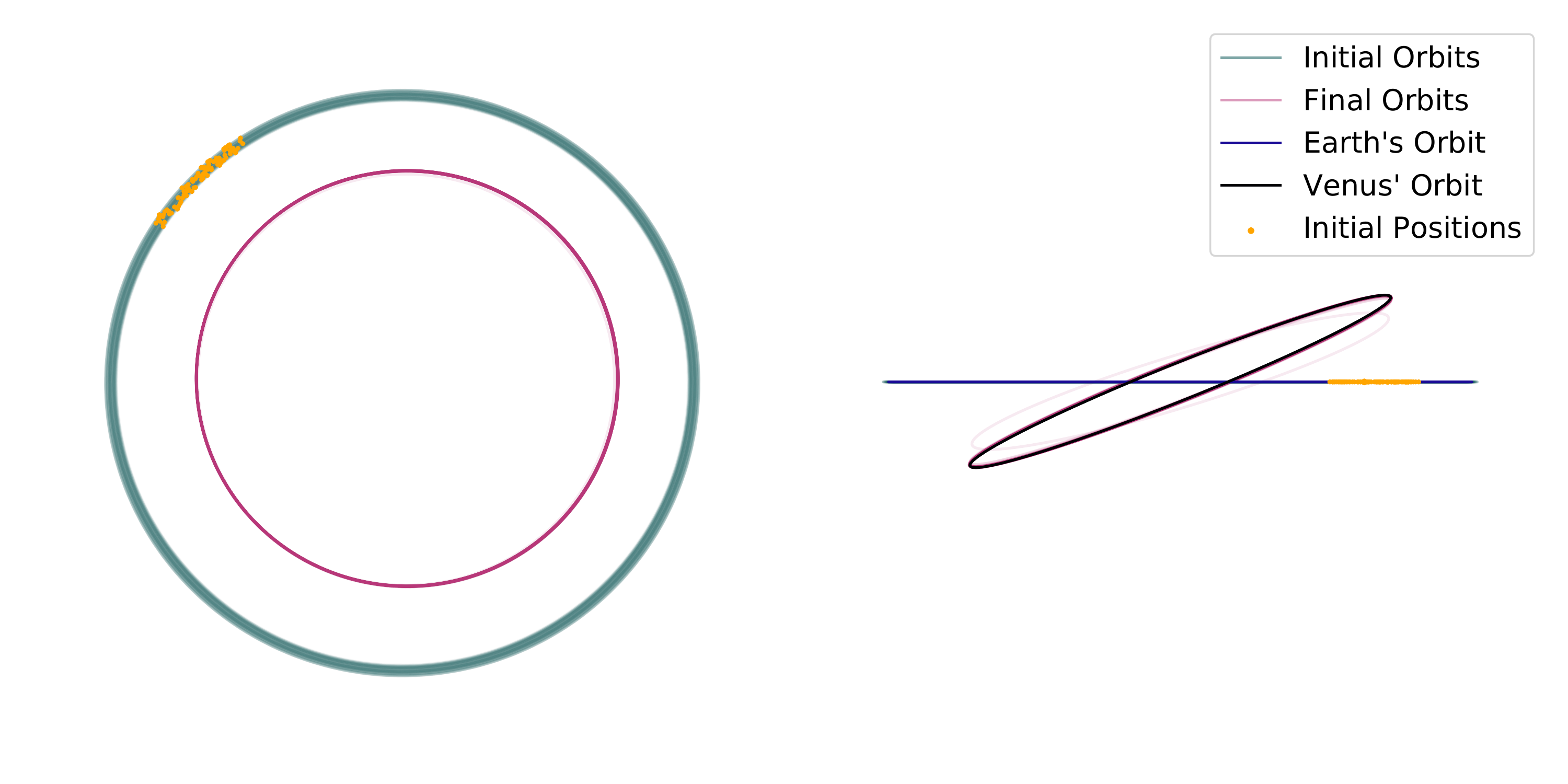}
    }
    
    \subfloat[$\mathcal A_8$: perturbation of 8\%\label{fig:final_orbit_halos:vf_w_grad:8p}]{
        \centering
        \includegraphics[width=0.80\columnwidth]{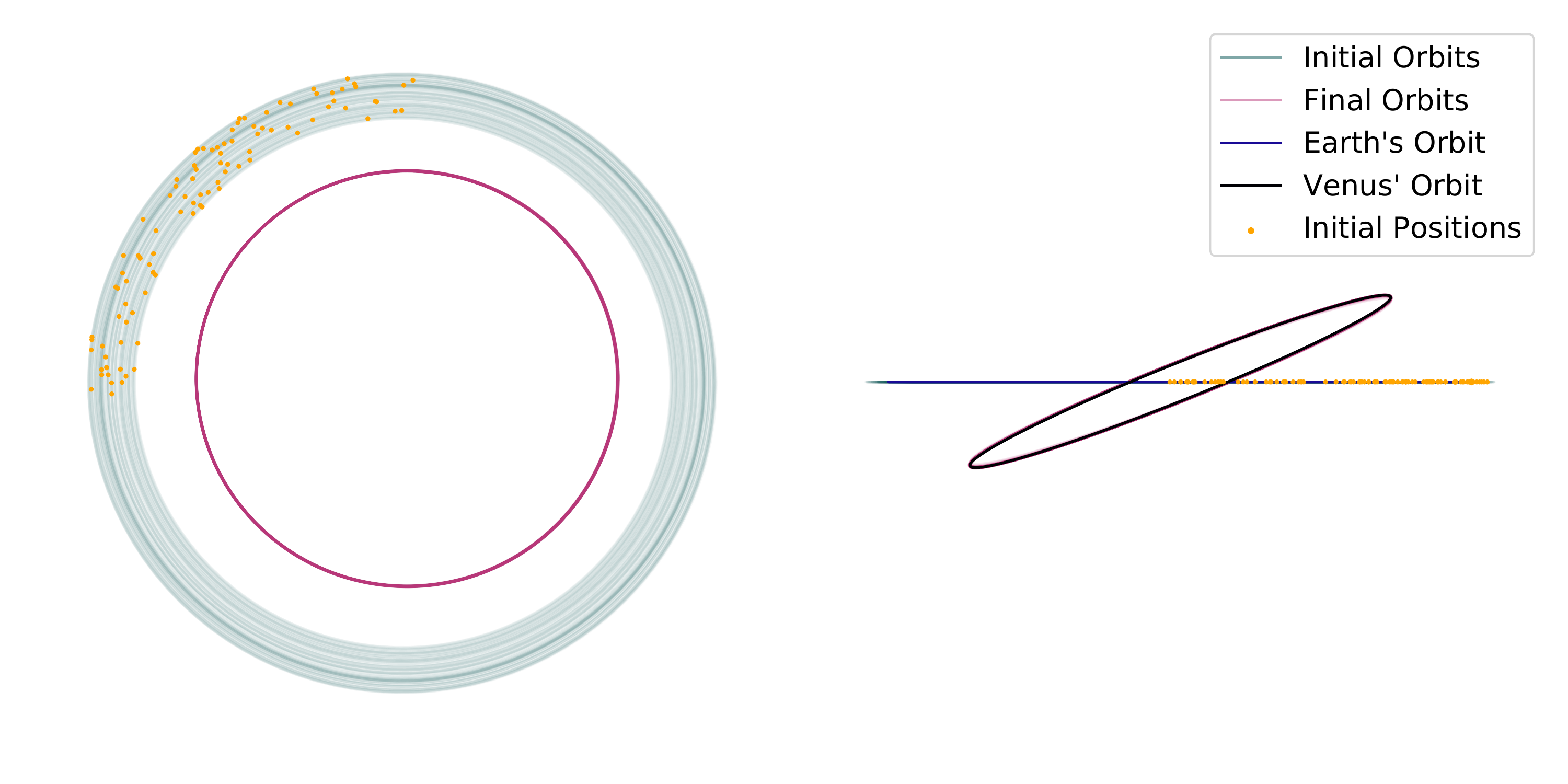}
    }
    
    \subfloat[$\mathcal A_{16}$: perturbation of 16\%\label{fig:final_orbit_halos:vf_w_grad:16p}]{
        \centering
        \includegraphics[width=0.80\columnwidth]{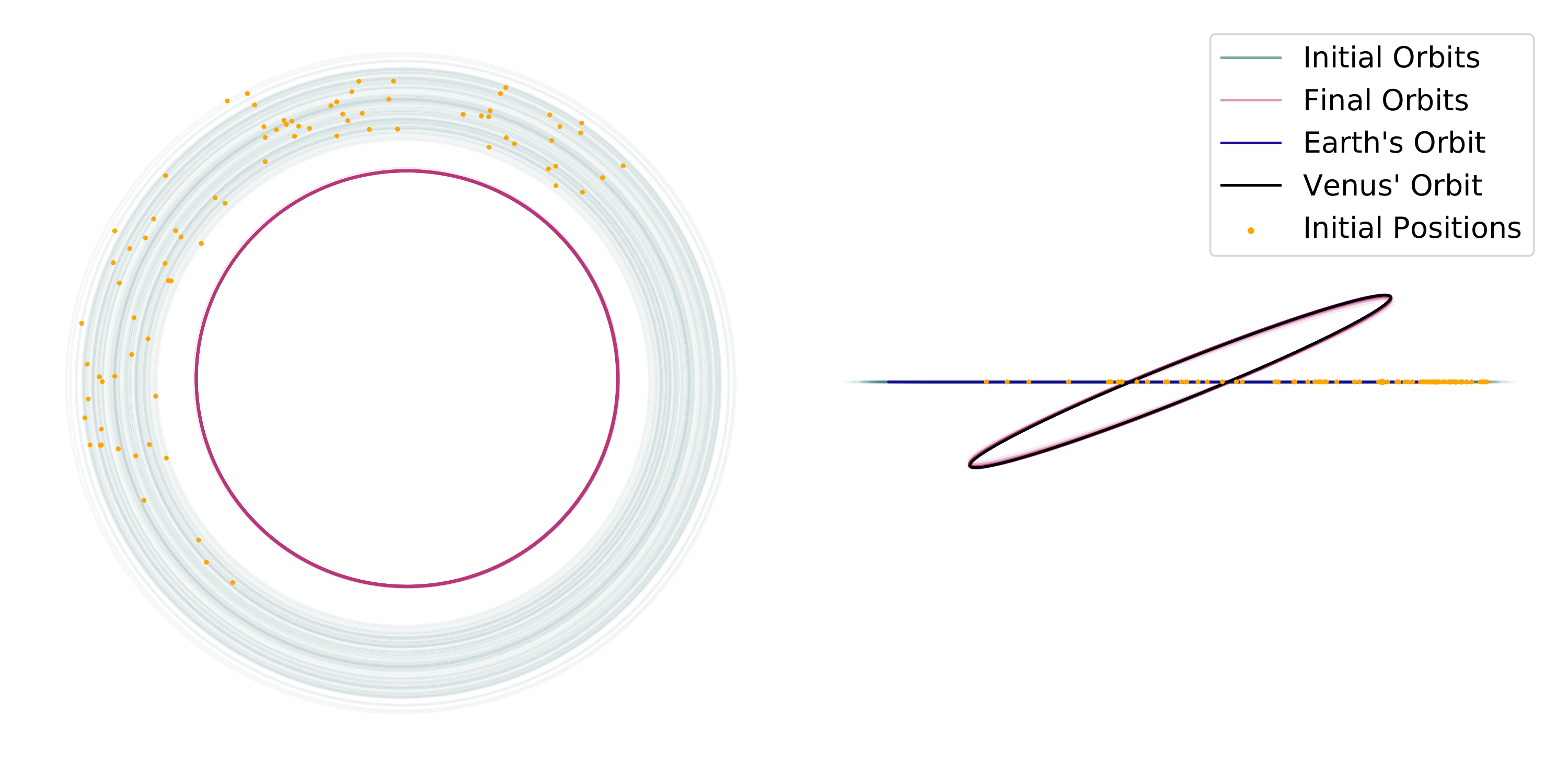}
    }
    
    \caption{Initial and final orbits of successful optimal transfers driven by $\mathcal{N}_3$ trained on database \emph{D}.}
    \label{fig:final_orbit_halos:vf_w_grad}
\end{figure}

Table \ref{table:performance_on_nominal_perturbed} we show the mean rEd and the success rates of the G\&CNETs when the spacecraft starts from initial conditions sampled at random in 4 different regions of increasing size centered around the nominal initial state ($\mathcal{A}_{2}$, $\mathcal{A}_{4}$, $\mathcal{A}_{8}$, $\mathcal{A}_{16}$).
A transfer is considered as successful when the minimum rEd reached along a transfer is below a threshold of $0.01$.
The regions are defined by perturbing the equinoctial elements of the initial nominal state (Earth's orbit) by $x\%$ for $\mathcal{A}_{x}$, e.g. for $2\%$ the initial states are perturbed to be between $98\%$ and $102\%$ of its original value.
The regions of these perturbations can be seen in Figure \ref{fig:final_orbit_halos:vf_w_grad} where we also visualize the corresponding final orbits for the case of the value function network $l_{\mathcal N_3}$ trained on database D.
The mean rEd reported in Table \ref{table:performance_on_nominal_perturbed} is computed considering the minimum rEd achieved along $N=100$ transfers starting from different initial conditions randomly sampled in a given region, in formal terms: $\mean{\mbox{rEd}} =\frac{1}{N}\sum_i{\min\limits_t{\norm{\Delta(\mathbf x(t))}{2}}}$.

\begin{table}[ht!]
\sisetup{group-separator={,},
         output-decimal-marker={.},
         group-four-digits,
         per-mode=fraction,
         group-digits=integer,
         round-mode=figures,
         round-precision=3,
         separate-uncertainty=false}
    \begin{center}
    \renewcommand{\arraystretch}{0.8}
    \setlength{\tabcolsep}{10.0pt}
         \begin{tabular}{@{}cllcccc@{}}
         \toprule
         & Training Database           &  &  & \multicolumn{1}{c}{\emph{A}} & \multicolumn{1}{c}{\emph{D}} & \multicolumn{1}{c}{\emph{G}} \\
         \cmidrule{2-7}
         
         \parbox[t]{3mm}{\multirow{16}{*}{\rotatebox[origin=c]{90}{NN Architectures}}}
         & \multirow{8}{*}{\emph{policy network} {} $l_{\mathcal{N}_1}$} & \multirow{2}{*}{$\mathcal{A}_{2}$} & $\mean{rEd}$
         & 0.0015(5) & 0.0063(3) & 0.0019(7) \\
         & & & Success Rate (\%)
         & 100.0 & 100.0 & 100.0 \\
         & & \multirow{2}{*}{$\mathcal{A}_{4}$} & $\mean{rEd}$
         & 0.0028(9) & 0.006(6) & 0.004(2) \\
         & & & Success Rate (\%)
         & 100.0 & 100.0 & 100.0 \\
         & & \multirow{2}{*}{$\mathcal{A}_{8}$} & $\mean{rEd}$
         & 0.005(3) & 0.0061(9) & 0.007(4) \\
         & & & Success Rate (\%)
         & 94.0 & 100.0 & 79.0 \\
         & & \multirow{2}{*}{$\mathcal{A}_{16}$} & $\mean{rEd}$
         & 0.011(8) & 0.007(2) & 0.013(0) \\
         & & & Success Rate (\%)
         & 51.0 & 96.0 & 46.0 \\
         
         \cmidrule{2-7}
         \cmidrule{2-7}
         & \multirow{8}{*}{\emph{value function network} {} $l_{\mathcal{N}_3}$} & \multirow{2}{*}{$\mathcal{A}_{2}$} & $\mean{rEd}$
         & 0.004(2) & 0.0013(7) & 0.0013(5) \\
         & & & Success Rate (\%)
         & 100.0 & 100.0 & 100.0 \\
         & & \multirow{2}{*}{$\mathcal{A}_{4}$} & $\mean{rEd}$
         & 0.006(3) & 0.0013(7) & 0.003(1) \\
         & & & Success Rate (\%)
         & 100.0 & 100.0 & 100.0 \\
         & & \multirow{2}{*}{$\mathcal{A}_{8}$} & $\mean{rEd}$
         & 0.011(8) & 0.01(3) & 0.005(3) \\
         & & & Success Rate (\%)
         & 65.0 & 98.0 & 99.0 \\
         & & \multirow{2}{*}{$\mathcal{A}_{16}$} & $\mean{rEd}$
         & 0.03(3) & 0.02(6) & 0.01(1) \\
         & & & Success Rate (\%)
         & 23.0 & 83.0 & 60.0 \\
        \bottomrule
        \end{tabular}
    \end{center}
    \caption{The mean rEd (upper line) and the success rate (lower line) of each network.}
    \label{table:performance_on_nominal_perturbed}
\end{table}

In Table \ref{table:performance_on_nominal_perturbed} we report in details the performances of the architectures $\mathcal{N}_1$ and $\mathcal{N}_3$ showing their high success rate and precision also far away from the nominal transfer The mean rEd values have a small standard deviation and are below a value of $0.05$ even in the worst case scenarios. We note that increasing the database size and density (i.e. changing database from from \emph{A} to \emph{D}) translates to an improved performance in all the perturbation regions at the cost of an increase in the mean rEd for the \emph{policy networks} for perturbations less than $16\%$. In terms of the \emph{policy networks}, if your goal is to fly to Venus' orbit successfully within a large region the best performing \emph{policy network} is the one trained on database \emph{D}. Conversely, if your goal is to acquire the target Venus orbit with high precision starting from initial conditions close to Earth, then the \emph{policy network} trained on database \emph{A} would be more suitable given the smaller mean rEd.
On the other hand, the \emph{value function network} show a significant improvement in the mean rEd and success rate when moving from database \emph{A} to \emph{D}. In the case of region $\mathcal{A}_{16}$, we see a deterioration of the performance of the \emph{value function network} when comparing database \emph{D} to database \emph{G}. It is difficult to pick one \emph{value function network} as being suitable for the case of a large perturbation given that the mean rEd of the network trained on database \emph{G} is half of that of the network trained on database \emph{D}, but fails to achieve Venus' orbit (within an rEd of $0.01$) $40\%$ of the time as opposed to $17\%$ for the training on database \emph{D}. For a small perturbation region, it is much easier to select the appropriate network given that the \emph{value function network} trained on database \emph{G} has the smallest rEd and smallest spread in rEd while achieving a $100\%$ success rate.

It is also curious to note that all the networks in Table \ref{table:performance_on_nominal_perturbed} are $100\%$ successful in achieving Venus' orbit within a rEd of $0.01$ for the regions $\mathcal{A}_2$ and $\mathcal{A}_4$. Furthermore, excluding the \emph{value function network} trained on database \emph{A} and the \emph{policy network} trained on database \emph{G}, the networks are able to achieve Venus' orbit with a very high success rate also in the region $\mathcal{A}_8$. This is very surprising considering that, as seen in Figure \ref{fig:final_orbit_halos:vf_w_grad}, region $\mathcal{A}_8$ is very large in terms of coverage.

\section{Conclusion}
\label{sec:conclusion}

In this work we have introduced a new methodology called \lq\lq backward generation of optimal examples\rq\rq\ to generate large databases of optimal trajectories bypassing entirely all the difficulties associated to optimal control solvers. In the case of a mass optimal interplanetary transfer between the Earth and Venus orbit we demonstrate the method building several databases of mass optimal transfer containing 4,000,000 optimal trajectories, largely surpassing any other previous related attempt. 

We find that deep artificial neural networks can learn the optimal policy (optimal thrust magnitude and direction) from these large databases both when predicting directly the optimal policy (policy network) and when predicting the value function (value function networks). In this last case we find that it is necessary to include some additional component to the loss function to inform its gradients in order for the approximated model to be used to recover the optimal policy.

The best performing networks (G\&CNETs) trained in this work are able to predict the optimal thrust and thrust direction to within an error of $5\%$ and $\ang{1}$. The \emph{value function networks} are also able to predict the optimal propellant mass to within 1\% of the true optimal mass. When starting from nominal initial condition the developed G\&CNET is able to complete the interplanetary transfer using only 2\permil\ more propellant than in the corresponding mathematical optimal solution. Furthermore, we find that the trained G\&CNETs are able to steer optimally the spacecraft to Venus' orbit also when deviating consistently from the planned nominal conditions.
Our results constitute a step forward towards the realization of a purely onboard system able to perform the guidance and control functions of a low-thrust spacecraft simultaneously and in real time.

%
%

\newpage
\clearpage

\section*{Appendix}
\label{sec:app}
This appendix contains the explicit forms of all the derivatives necessary to write Eq.~\eqref{eq:eom_costates} explicitly.
\subsection*{The $\dot\lambda_p$ equation}
We get:
\begin{align}
\dot\lambda_p & = - \frac {c_1 u} m \boldsymbol\lambda^T \frac{\partial \mathbf B(\mathbf x)}{\partial p}  \mathbf i_\tau -  w^2 \lambda_L \frac{\partial}{\partial p} \sqrt{\frac{\mu}{p^3}} \\ &
=  - \frac {c_1 u} m \boldsymbol\lambda^T \frac{\partial \mathbf B(\mathbf x)}{\partial p}  \mathbf i_\tau  +\frac 32  w^2 \lambda_L \sqrt{\frac \mu{p^5}}  \nonumber
\end{align}
\begin{equation}
 2 \sqrt{\mu p} \frac{\partial \mathbf B(\mathbf x)}{\partial p} =
\left[
\begin{array}{ccc}
0 &  \frac {6p}w  & 0 \\
 \sin L & [(1+w)\cos L + f]\frac 1w  & - \frac gw (h\sin L-k\cos L)  \\
- \cos L & [(1+w)\sin L + g]\frac 1w  & \frac fw (h\sin L-k\cos L)  \\
0 & 0  & \frac 1w \frac{s^2}{2}\cos L \\
0 & 0  & \frac 1w \frac{s^2}{2}\sin L \\
0 & 0  & \frac 1w (h\sin L - k\cos L) \\
\end{array}
\right]
\end{equation}

\subsection*{The $\dot\lambda_f$ equation}
We get:
\begin{align}
\dot\lambda_f & = - \frac {c_1 u} m \boldsymbol\lambda^T \frac{\partial \mathbf B(\mathbf x)}{\partial f}  \mathbf i_\tau - 2 \lambda_L w \sqrt{\frac{\mu}{p^3}}  \frac{\partial w}{\partial f}  \\ &
= - \frac {c_1 u} m \boldsymbol\lambda^T \frac{\partial \mathbf B(\mathbf x)}{\partial f}  \mathbf i_\tau - 2 \lambda_L w \sqrt{\frac{\mu}{p^3}} \cos L   \nonumber
\end{align}

\begin{equation}
 w^2 \sqrt{\frac \mu p} \frac{\partial \mathbf B(\mathbf x)}{\partial f} =
\left[
\begin{array}{ccc}
0 & - 2p \cos L  & 0 \\
0 & w - (\cos L + f)\cos L & g\cos L (h\sin L-k\cos L)    \\
0 & -(\sin L + g)\cos L &  (w-f\cos L) (h\sin L-k\cos L)  \\
0 & 0  & - \frac{s^2}{2}\cos^2 L \\
0 & 0  & -  \frac{s^2}{2}\sin L\cos L \\
0 & 0  & - (h\sin L - k\cos L) \cos L\\
\end{array}
\right]
\end{equation}

\subsection*{The $\dot\lambda_g$ equation}
We get:
\begin{align}
\dot\lambda_g & = - \frac {c_1 u} m \boldsymbol\lambda^T \frac{\partial \mathbf B(\mathbf x)}{\partial g}  \mathbf i_\tau - 2 \lambda_L w \sqrt{\frac{\mu}{p^3}}  \frac{\partial w}{\partial g}  \\ &
= - \frac {c_1 u} m \boldsymbol\lambda^T \frac{\partial \mathbf B(\mathbf x)}{\partial g}  \mathbf i_\tau - 2 \lambda_L w \sqrt{\frac{\mu}{p^3}} \sin L   \nonumber
\end{align}

\begin{equation}
 w^2 \sqrt{\frac \mu p} \frac{\partial \mathbf B(\mathbf x)}{\partial g} =
\left[
\begin{array}{ccc}
0 & - {2p} \sin L  & 0 \\
0 & - (\cos L + f) \sin L & - (w-g\sin L) (h\sin L-k\cos L)  \\
0 & w - (\sin L + g)\sin L & - f \sin L (h\sin L-k\cos L)   \\
0 & 0  & - \frac{s^2}{2}\cos L\sin L \\
0 & 0  & - \frac{s^2}{2}\sin^2 L \\
0 & 0  & -(h\sin L - k\cos L) \sin L\\
\end{array}
\right]
\end{equation}

\subsection*{The $\dot\lambda_h$ equation}
We get:
\begin{align}
\dot\lambda_h & = - \frac {c_1 u} m \boldsymbol\lambda^T \frac{\partial \mathbf B(\mathbf x)}{\partial h}  \mathbf i_\tau
\end{align}

$$
\frac{\partial \mathbf B(\mathbf x)}{\partial h} =  \sqrt{\frac p \mu}
\left[
\begin{array}{ccc}
0 & 0 & 0 \\
0 & 0  & - \frac gw \sin L  \\
0 & 0  & \frac fw \sin L  \\
0 & 0  & \frac hw \cos L \\
0 & 0  & \frac hw \sin L \\
0 & 0  & \frac 1w \sin L \\
\end{array}
\right]
$$

\subsection*{The $\dot\lambda_k$ equation}
We get:
\begin{equation}
\dot\lambda_k = - \frac {c_1 u} m \boldsymbol\lambda^T \frac{\partial \mathbf B(\mathbf x)}{\partial k}  \mathbf i_\tau
\end{equation}

\begin{equation}
\frac{\partial \mathbf B(\mathbf x)}{\partial k} =  \sqrt{\frac p \mu}
\left[
\begin{array}{ccc}
0 & 0  & 0 \\
0 & 0  & \frac gw\cos L  \\
0 & 0  & - \frac fw \cos L  \\
0 & 0  & \frac kw \cos L \\
0 & 0  & \frac kw \sin L \\
0 & 0  & -\frac 1w \cos L \\
\end{array}
\right]
\end{equation}

\subsection*{The $\dot \lambda_L$ equation}
We get:
\begin{equation}
\dot\lambda_L = - \frac {c_1 u} m \boldsymbol\lambda^T \frac{\partial \mathbf B(\mathbf x)}{\partial L}  \mathbf i_\tau - 2 w \sqrt{\frac{\mu}{p^3}} \lambda_L w_L
\end{equation}
where,

\begin{multline}
w^2 \sqrt{\frac \mu p} \frac{\partial \mathbf B(\mathbf x)}{\partial L} = 
\\ 
\\
\left[
\begin{array}{ccc}
0 & -2p (g\cos L - f\sin L) & 0 \\
w^2\cos L & - (1+w)w\sin L - w_L(\cos L + f)  & ((wh+w_Lk)\cos L + (wk-w_Lh)\sin L) g  \\
w^2\sin L & (1+w)w\cos L - w_L(\sin L + g)  & ((wh+w_Lk)\cos L + (wk-w_Lh)\sin L) f  \\
0 & 0  & -\frac{s^2}{2}(w\sin L + w_L\cos L) \\
0 & 0  &  \frac{s^2}{2}(w\cos L - w_L\sin L) \\
0 & 0  &  {(wh+w_Lk)\cos L + (wk-w_Lh)\sin L} \\
\end{array}
\right]
\end{multline}
where,
\begin{equation}
w_L = \frac{\partial w}{\partial L} = g\cos L - f\sin L
\end{equation}
\normalsize

\subsection*{The $\dot \lambda_m$ equation}
We get:
\begin{align}
\dot\lambda_m & = - \frac {c_1 u} {m^2} |\boldsymbol\lambda^T \mathbf B(\mathbf x)|
\end{align}



\bibliographystyle{aiaa}

\bibliography{main.bib}
\end{document}

%% file: gecnet_draw.tex
\begin{figure}[tb]
\centering
\def\layersep{2cm}
\def\nodesep{0.3cm}

\begin{tikzpicture}[shorten >=1pt,->,draw=black!50, node distance=\layersep, scale=0.7]
    \tikzstyle{every pin edge}=[<-,shorten <=1pt]
    \tikzstyle{neuron}=[circle,fill=black!25,minimum size=5pt,inner sep=0pt]
    \tikzstyle{input neuron}=[neuron, fill=gray!50];
    \tikzstyle{output neuron}=[neuron, fill=red!50];
    \tikzstyle{output neuron2}=[neuron, fill=green!50];

    \tikzstyle{hidden neuron softplus}=[neuron, fill=blue!50];

    \tikzstyle{annot} = [text width=4em, text centered]

    \node[input neuron, pin=left:$p$] (I-1) at (0,-\nodesep*1*1.63) {};
    \node[input neuron, pin=left:$f$] (I-2) at (0,-\nodesep*2*1.63) {};
    \node[input neuron, pin=left:$g$] (I-3) at (0,-\nodesep*3*1.63) {};
    \node[input neuron, pin=left:$h$] (I-4) at (0,-\nodesep*4*1.63) {};
    \node[input neuron, pin=left:$k$] (I-5) at (0,-\nodesep*5*1.63) {};
    \node[input neuron, pin=left:$L$] (I-6) at (0,-\nodesep*6*1.63) {};
    \node[input neuron, pin=left:$m$] (I-7) at (0,-\nodesep*7*1.63) {};

    \node[] (L0) at (0,-\nodesep*7*1.63-4.6*\nodesep) {$L^{(0)}$};

    \foreach \name / \y in {1,...,15}
        \path[yshift=0.5cm]
            node[hidden neuron softplus] (H1-\name) at (\layersep,-\nodesep*\y) {};
    \node[] (L1) at (\layersep,-\nodesep*15-1*\nodesep) {$L^{(1)}$};

    \foreach \name / \y in {1,...,15}
        \path[yshift=0.5cm]
            node[hidden neuron softplus] (H2-\name) at (2*\layersep,-\nodesep*\y) {};
    \node[] (L2) at (2*\layersep,-\nodesep*15-1*\nodesep) {$L^{(2)}$};
            
    \foreach \name / \y in {1,...,15}
        \path[yshift=0.5cm]
            node[hidden neuron softplus] (H3-\name) at (2.5*\layersep,-\nodesep*\y) {};
    \node[] (Ll) at (2.5*\layersep,-\nodesep*15-1*\nodesep) {$L^{(l)}$};

    \node[output neuron2, pin={[pin edge={->}]right:$v$}, right of=H3-3] (O-1) at (2.3*\layersep,-1.8)  {};
    \node[below of=O-2] (Ll1) at (3.8*\layersep,-6.6*\nodesep) {$L^{(l+1)}$};

    \foreach \source in {1,...,7}
        \foreach \dest in {1,...,15}
            \path (I-\source) edge (H1-\dest);
            
    \foreach \source in {1,...,15}
        \foreach \dest in {1,...,15}
            \path (H1-\source) edge (H2-\dest);

    \path[thick] (H2-8) edge[-,loosely dotted] (H3-8);

    \foreach \source in {1,...,15}
        \foreach \dest in {1,...,1}
            \path (H3-\source) edge (O-\dest);

    \pgfmathsetmacro{\scale}{0.2}
    \pgfmathsetmacro{\Ox}{10}
    \pgfmathsetmacro{\Oy}{4}

    \draw[scale=\scale, shift = {(\Ox, \Oy)}] (-3,0) -- (3,0) node[right] {};
    \draw[scale=\scale, shift = {(\Ox, \Oy)}] (0,-1) -- (0,3) node[above] {softplus};
    \draw[scale=\scale,domain=-3:3,smooth,variable=\x,blue, shift = {(\Ox, \Oy)} ] plot ({\x},{ln(1+exp(\x)});
    
    \draw[scale=\scale, shift = {(\Ox+10, \Oy)}] (-3,0) -- (3,0) node[right] {};
    \draw[scale=\scale, shift = {(\Ox+10, \Oy)}] (0,-1) -- (0,3) node[above] {};
    \draw[scale=\scale,domain=-3:3,smooth,variable=\x,blue, shift = {(\Ox+10, \Oy)} ] plot ({\x},{ln(1+exp(\x)});
    
    \draw[scale=\scale, shift = {(\Ox+15, \Oy)}] (-1,0) -- (3,0) node[right] {};
    \draw[scale=\scale, shift = {(\Ox+15, \Oy)}] (0,-1) -- (0,3) node[above] {};
    \draw[scale=\scale,domain=-2:3,smooth,variable=\x,blue, shift = {(\Ox+15, \Oy)} ] plot ({\x},{ln(1+exp(\x)});
    
    \draw[scale=\scale, shift = {(\Ox+25, \Oy-2)}] (-1,0) -- (3,0) node[right] {};
    \draw[scale=\scale, shift = {(\Ox+25, \Oy-2)}] (0,-1) -- (0,3) node[above] {linear};
    \draw[scale=\scale,domain=-2:3,smooth,variable=\x,green, shift = {(\Ox+25, \Oy-2)} ] plot ({\x},{(\x)});

\end{tikzpicture}
\begin{tikzpicture}[shorten >=1pt,->,draw=black!50, node distance=\layersep, scale=0.7]
    \tikzstyle{every pin edge}=[<-,shorten <=1pt]
    \tikzstyle{neuron}=[circle,fill=black!25,minimum size=5pt,inner sep=0pt]
    \tikzstyle{input neuron}=[neuron, fill=gray!50];
    \tikzstyle{output neuron}=[neuron, fill=red!50];
    \tikzstyle{output neuron2}=[neuron, fill=green!50];

    \tikzstyle{hidden neuron softplus}=[neuron, fill=blue!50];

    \tikzstyle{annot} = [text width=4em, text centered]

    \node[input neuron, pin=left:$p$] (I-1) at (0,-\nodesep*1*1.63) {};
    \node[input neuron, pin=left:$f$] (I-2) at (0,-\nodesep*2*1.63) {};
    \node[input neuron, pin=left:$g$] (I-3) at (0,-\nodesep*3*1.63) {};
    \node[input neuron, pin=left:$h$] (I-4) at (0,-\nodesep*4*1.63) {};
    \node[input neuron, pin=left:$k$] (I-5) at (0,-\nodesep*5*1.63) {};
    \node[input neuron, pin=left:$L$] (I-6) at (0,-\nodesep*6*1.63) {};
    \node[input neuron, pin=left:$m$] (I-7) at (0,-\nodesep*7*1.63) {};

    \node[] (L0) at (0,-\nodesep*7*1.63-4.6*\nodesep) {$L^{(0)}$};

    \foreach \name / \y in {1,...,15}
        \path[yshift=0.5cm]
            node[hidden neuron softplus] (H1-\name) at (\layersep,-\nodesep*\y) {};
    \node[] (L1) at (\layersep,-\nodesep*15-1*\nodesep) {$L^{(1)}$};

    \foreach \name / \y in {1,...,15}
        \path[yshift=0.5cm]
            node[hidden neuron softplus] (H2-\name) at (2*\layersep,-\nodesep*\y) {};
    \node[] (L2) at (2*\layersep,-\nodesep*15-1*\nodesep) {$L^{(2)}$};
            
    \foreach \name / \y in {1,...,15}
        \path[yshift=0.5cm]
            node[hidden neuron softplus] (H3-\name) at (2.5*\layersep,-\nodesep*\y) {};
    \node[] (Ll) at (2.5*\layersep,-\nodesep*15-1*\nodesep) {$L^{(l)}$};

    \node[output neuron, pin={[pin edge={->}]right:$u$}, right of=H3-3] (O-1) at (2.3*\layersep,-0.2)  {};
    \node[output neuron2, pin={[pin edge={->}]right:$d_r$}, right of=H3-3] (O-2) at (2.3*\layersep,-3)  {};
    \node[output neuron2, pin={[pin edge={->}]right:$d_t$}, right of=H3-3] (O-3) at (2.3*\layersep,-3.4)  {};
    \node[output neuron2, pin={[pin edge={->}]right:$d_n$}, right of=H3-3] (O-4) at (2.3*\layersep,-3.8)  {};
    \node[below of=O-2] (Ll1) at (3.8*\layersep,-6.6*\nodesep) {$L^{(l+1)}$};

    \foreach \source in {1,...,7}
        \foreach \dest in {1,...,15}
            \path (I-\source) edge (H1-\dest);
            
    \foreach \source in {1,...,15}
        \foreach \dest in {1,...,15}
            \path (H1-\source) edge (H2-\dest);

    \path[thick] (H2-8) edge[-,loosely dotted] (H3-8);

    \foreach \source in {1,...,15}
        \foreach \dest in {1,...,4}
            \path (H3-\source) edge (O-\dest);

    \pgfmathsetmacro{\scale}{0.2}
    \pgfmathsetmacro{\Ox}{10}
    \pgfmathsetmacro{\Oy}{4}

    \draw[scale=\scale, shift = {(\Ox, \Oy)}] (-3,0) -- (3,0) node[right] {};
    \draw[scale=\scale, shift = {(\Ox, \Oy)}] (0,-1) -- (0,3) node[above] {softplus};
    \draw[scale=\scale,domain=-3:3,smooth,variable=\x,blue, shift = {(\Ox, \Oy)} ] plot ({\x},{ln(1+exp(\x)});
    
    \draw[scale=\scale, shift = {(\Ox+10, \Oy)}] (-3,0) -- (3,0) node[right] {};
    \draw[scale=\scale, shift = {(\Ox+10, \Oy)}] (0,-1) -- (0,3) node[above] {};
    \draw[scale=\scale,domain=-3:3,smooth,variable=\x,blue, shift = {(\Ox+10, \Oy)} ] plot ({\x},{ln(1+exp(\x)});
    
    \draw[scale=\scale, shift = {(\Ox+15, \Oy)}] (-1,0) -- (3,0) node[right] {};
    \draw[scale=\scale, shift = {(\Ox+15, \Oy)}] (0,-1) -- (0,3) node[above] {};
    \draw[scale=\scale,domain=-2:3,smooth,variable=\x,blue, shift = {(\Ox+15, \Oy)} ] plot ({\x},{ln(1+exp(\x)});
    
    \draw[scale=\scale, shift = {(\Ox+30, \Oy-2)}] (-1,0) -- (3,0) node[right] {};
    \draw[scale=\scale, shift = {(\Ox+30, \Oy-2)}] (0,-1) -- (0,3) node[above] {sigmoid};
    \draw[scale=\scale,domain=-2:3,smooth,variable=\x,red, shift = {(\Ox+30, \Oy-2)} ] plot ({\x},{1/(1+exp(-\x * 2))});
    
    \draw[scale=\scale, shift = {(\Ox+30, \Oy-15)}] (-1,0) -- (3,0) node[right] {};
    \draw[scale=\scale, shift = {(\Ox+30, \Oy-14)}] (0,-1) -- (0,3) node[above] {linear};
    \draw[scale=\scale,domain=-2:3,smooth,variable=\x,green, shift = {(\Ox+30, \Oy-15)} ] plot ({\x},{(\x)});
    
\end{tikzpicture}
\caption{Value function network (left) and policy network (right). \label{fig:ffnn}}

\end{figure}